\documentclass[10pt,twocolumn,letterpaper]{article}

\usepackage{cvpr}
\usepackage{times}
\usepackage{epsfig}
\usepackage{graphicx}
\usepackage{amsmath}
\usepackage{amssymb}

\usepackage{booktabs}
\usepackage{multirow}
\usepackage{threeparttable}
\usepackage{bm}
\usepackage{setspace}
\usepackage{pdfpages}

\usepackage[breaklinks=true,bookmarks=false]{hyperref}

\cvprfinalcopy 

\def\cvprPaperID{3654} 

\begin{document}

\title{LO-Net: Deep Real-time Lidar Odometry}

\author{
Qing Li$^1$, Shaoyang Chen$^1$, Cheng Wang$^1$*, Xin Li$^2$, Chenglu Wen$^1$, Ming Cheng$^1$, Jonathan Li$^1$\\
$^1$Xiamen University, Fujian, China\\ 
$^2$Louisiana State University, Louisiana, USA\\
}

\maketitle


\begin{abstract}
We present a novel deep convolutional network 
pipeline, LO-Net, for real-time lidar odometry estimation. 
Unlike most existing lidar odometry (LO) estimations that go through individually designed feature selection, feature matching, 
and pose estimation pipeline, LO-Net can be trained in an end-to-end manner. 
With a new mask-weighted geometric constraint loss, LO-Net can effectively learn feature representation for LO estimation, 
and can implicitly exploit the sequential dependencies and dynamics in the data.
We also design a scan-to-map module, which uses the geometric and semantic information learned in LO-Net, 
to improve the estimation accuracy. 
Experiments on benchmark datasets demonstrate that LO-Net outperforms existing learning based approaches 
and has similar accuracy with the state-of-the-art geometry-based approach, LOAM.

\end{abstract}

\section{Introduction}

Estimating 3D position and orientation of a mobile platform is a fundamental problem in 3D computer vision, 
and it provides important navigation information for robotics and autonomous driving. 
Mobile platforms usually collect information from the real-time perception of the environment and use on-board sensors 
such as lidars, Inertial Measurement Units (IMU), or cameras, to estimate their motions. 
Lidar can obtain robust features of different environments as it is not sensitive to lighting conditions, 
and it also acquires more accurate distance information than cameras. 
Therefore, developing an accurate and robust real-time lidar odometry estimation system is desirable. 

Classic lidar-based registration methods used in pose estimation include Iterative Closest Point (ICP)~\cite{paul1992method}, ICP variants~\cite{pomerleau2013comparing}, and feature-based approaches~\cite{rusu2009fast}. 
But due to the nonuniformity and sparsity of the lidar point clouds, 
these methods often fail to match such data.  
ICP approaches find transformations between consecutive scans by minimizing distances 
between corresponding points from these scans,  
but points in one frame may miss their spatial counterparts in the next frame, due to sparsity of scan resolution. 
Feature-based methods are less sensitive to the quality of scans, and hence, are more powerful. 
However, they are usually more computationally expensive. 
Furthermore, most feature-based methods are sensitive to another environmental factor, dynamic objects in the scene. 
These two issues inhibit many feature-based methods from producing effective odometry estimation.

\begin{figure}[t]
	\begin{center}
		\includegraphics[width=\linewidth]{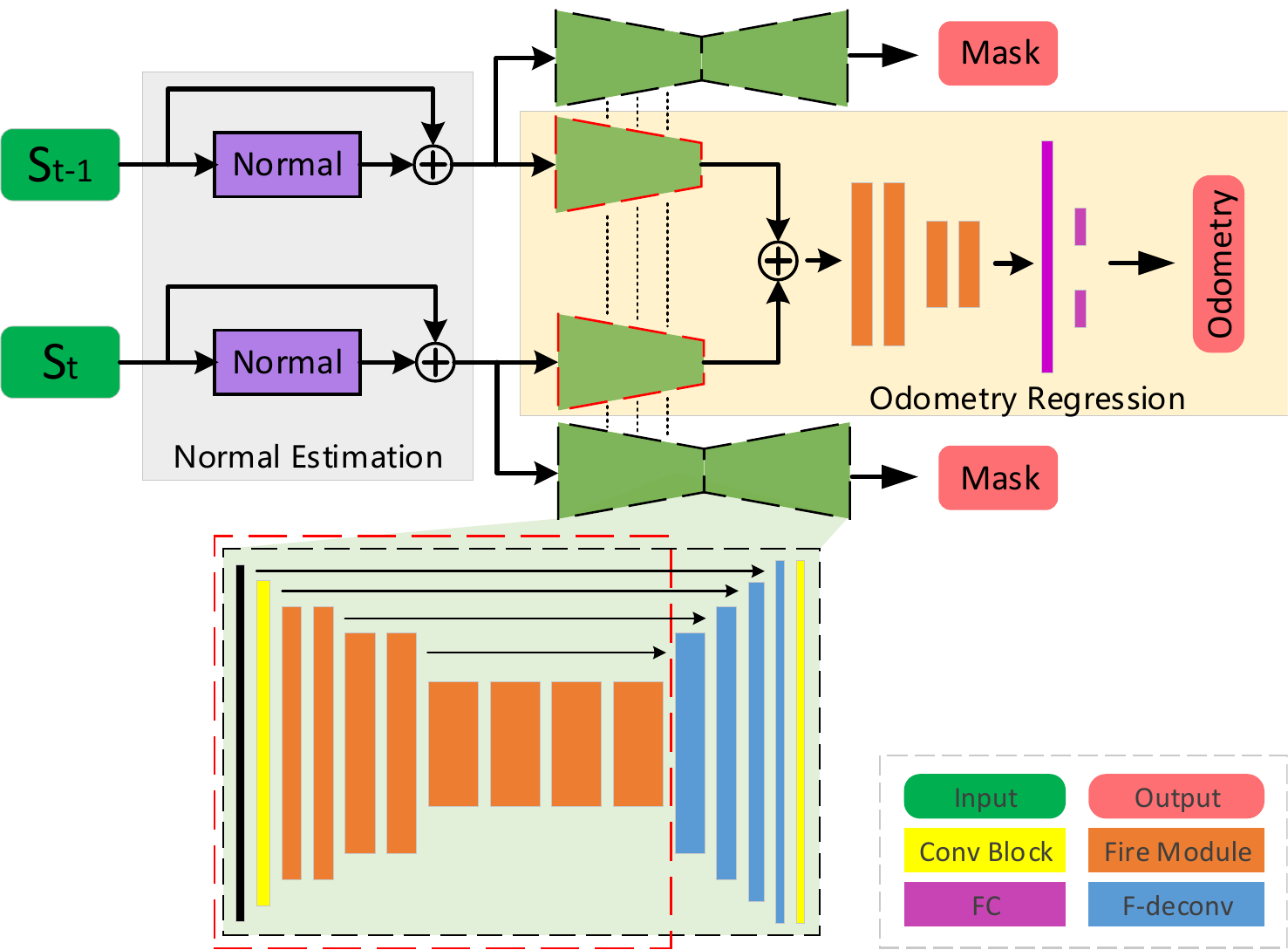}
	\end{center}
	\caption{Top: Data stream of LO-Net. 
		Bottom: Network architecture of feature extraction layers (red dashed line) and mask prediction layers (black dashed line) . 
		Our network takes two consecutive lidar scans as input and infers the relative 6-DoF pose. 
		The output data will be further processed by the mapping module.}
	\label{fig:net}
\end{figure}

Recently, deep learning-based methods have outperformed classic approaches in many computer vision problems. 
Many Convolutional Neural Networks (CNNs) architectures and training models have become the state-of-the-art in these tasks.  
However, the exploration of effective CNNs in some 3D geometric data processing problems, such as 6-DoF pose estimation, has not been this successful yet.  
Although quite a few CNN-based 6-DoF pose estimation (from RGB images) strategies~\cite{wang2017deepvo, zhou2017unsupervised, yang2018_dvso, yin2018geonet} have been explored recently,
these methods often suffer from the inaccurate depth prediction and scale drift. 
This paper proposes a new deep neural network architecture for lidar odometry estimation. 
We are inspired by the recent CNNs-based camera localization 
and pose regression works~\cite{zhou2017unsupervised,  Valada18icra, kendall2017geometric, yang2018lego} 
in the context of network structure design, 
and the traditional lidar odometry methods~\cite{zhang2014loam, moosmann2011velodyne, deschaud2018imls} in the aspect of lidar mapping.
The pipeline is illustrated in Figure~\ref{fig:net}. 
Our method can better accumulate motion-specific features by incorporating pairwise scans, 
better interpret the spatial relations of scans by applying normal consistency, 
and better locate effective static regions using a simultaneously learned uncertainty mask. 

The main contributions of this paper are as follows: 
1) We propose a novel scan-to-scan lidar odometry estimation network 
which simultaneously learns to estimate the normal and a mask for dynamic regions.
2) We incorporate a spatiotemporal geometry consistency constraint to the network,  
which provides higher-order interaction between consecutive scans and helps better regularize odometry learning.
3) We couple an efficient mapping module into the estimation pipeline. 
By utilizing the normal and mask information learned from LO-Net, we can achieve a still real-time but more accurate odometry estimation. 

We compared our approach with existing representative methods on  
commonly used benchmark datasets \emph{KITTI}~\cite{geiger2013vision, geiger2012we} and \emph{Ford Campus Vision and Lidar Data Set}~\cite{pandey2011ford}.
Our approach achieves the state-of-the-art results. 
To our best knowledge, this is the first neural network regression model that is comparable to the state-of-the-art geometry feature based techniques in 3D lidar odometry estimation.

\section{Related work}

\begin{spacing}{1.5}
\noindent
\textit{A. ICP variants}
\end{spacing}

Most existing lidar-based pose estimation algorithms are built upon variants of the ICP method~\cite{paul1992method, pomerleau2013comparing}. 
ICP iteratively matches the adjacent scans and estimates the pose transformation, 
by minimizing distances between corresponded elements, until a specific termination condition is met.
Despite its wide applicability, ICP is computationally expensive and sensitive to initial poses. 
Various ICP variants, such as point-to-plane or plane-to-plane ICP, were developed to improve ICP's convergence speed and robustness against local minima. 

By combining point-to-point ICP and point-to-plane ICP, Segal \etal~\cite{segal2009generalized} introduce a plane-to-plane strategy called Generalized ICP (GICP). 
The covariance matrices of the local surfaces is used to match corresponding point clouds. 
Their results show better accuracy over original ICP. 
The Normal Iterative Closest Point (NICP)~\cite{serafin2015nicp} takes into account the normal and curvature of each local surface point. 
Experiments show that NICP can offer better robustness and performance than the GICP.

Grant \etal\cite{grant2013finding} derive a novel Hough-based voting scheme for finding planes in laser point clouds. 
A plane-based registration method~\cite{pathak2010fast}, that aligns detected planes, is adopted to compute the final transformation. 
For the non-uniformly sampled point clouds from laser sensors, 
Serafin \etal~\cite{serafin2016fast} present a fast method to extract robust feature points. 
They remove ground points by a flat-region-removal method, then extract key-points of lines and planes, and use them to estimate the transformation. 
They show comparative results against the NARF key-point detector~\cite{steder2011point} 
and highlight that the feature extraction algorithm is applicable to the Simultaneous Localization and Mapping (SLAM) problem~\cite{hess2016real, cadena2016past}. 
Similarly, Douillard \etal~\cite{douillard2012scan} remove the ground points by voxelization, cluster the remaining points into the segments, 
then match these segments through a modified ICP algorithm. 

To overcome the sparsity of lidar point clouds, Collar Line Segments (CLS)~\cite{velas2016collar} groups the points into polar bins, and generates line segments within each bin. 
Then, ICP can be used to register the corresponding lines and estimate the transformation between two scans. 
Although it produces better results than GICP~\cite{segal2009generalized}, 
CLS is not real-time because the line segments computation is very slow. 

Over the past few years, Lidar Odometry And Mapping (LOAM)~\cite{zhang2014loam, zhang2017low} 
has been considered as the state-of-the-art lidar motion estimation method. 
It extracts the line and plane features in lidar data, 
and saves these features to the map for edge-line and plane-surface matching. 
LOAM dose not consider the dynamic objects in the scene and achieves low-drift and real-time odometry estimation by having two modules running in parallel.  
The estimated motion of scan-to-scan registration is used to correct the distortion of point clouds and guarantee the real-time performance. 
Finally, the odometry outputs are optimized through a map. 

\begin{spacing}{1.5}
	\noindent
	\textit{B. Deep learning-based methods}
\end{spacing}

Deep learning has achieved promising results on the issues of visual odometry (VO)~\cite{wang2017deepvo, zhou2017unsupervised,  zhan2018unsupervised, yang2018_dvso}, 
image-based pose estimation or localization~\cite{kendall2015posenet, kendall2017geometric, brahmbhatt2017mapnet}, 
and point cloud classification and segmentation~\cite{qi2017pointnet, li2018pointcnn}.
Unfortunately, using deep learning methods to solve 3D lidar odometry problems still remains challenging. 
Velas \etal~\cite{velas2018cnn} use CNNs to do lidar odometry estimation on lidar scan sequences.  
To train the CNNs, the original lidar data is transformed into dense matrix with three channels.
However, this model formulates the odometry estimation as a classification problem, rather than a numerical regression,  
and it only estimates translational parameters. 
Hence, is not competent for accurate 6-DoF parameters estimation. 
The method of~\cite{velas2018cnn} and ours are conceptually similar, 
but our network better handles uncertain dynamic regions through mask prediction, 
and more effectively uses spatiotemporal consistency to regularize the learning for stable estimations.

\section{Method}     
\label{sec:method}

Odometry estimation algorithm uses consecutive point clouds obtained by a moving robot as inputs. 
Our LO-Net can interpret and exploit more information by performing sequential learning rather than processing a single scan. 
And the features learned by LO-Net encapsulate geometric information for the LO problem. 
As shown in Figure~\ref{fig:net}, our LO-Net architecture consists of a normal estimation sub-network, 
a mask prediction sub-network, and a Siamese pose regression main-network. 
LO-Net takes two consecutive scans $(S_{t-1};S_t)$ as input and jointly estimates the 6-DoF relative pose between the scans, 
point-wise normal vector, and a mask of moving objects for each scan. 
Despite being jointly trained, the three modules can be used independently for inference. 
The odometry outputs of the LO-Net are then refined through a lidar mapping, 
which registers the lidar point clouds onto a globally constructed map. 
The final output is the transformation of scan $S_t$ with respect to the initial reference position, 
namely, the pose of each $S_t$ in the world coordinate system.

\subsection{Lidar data encoding}

\begin{figure}[t]
	\begin{center}
		\includegraphics[width=\linewidth]{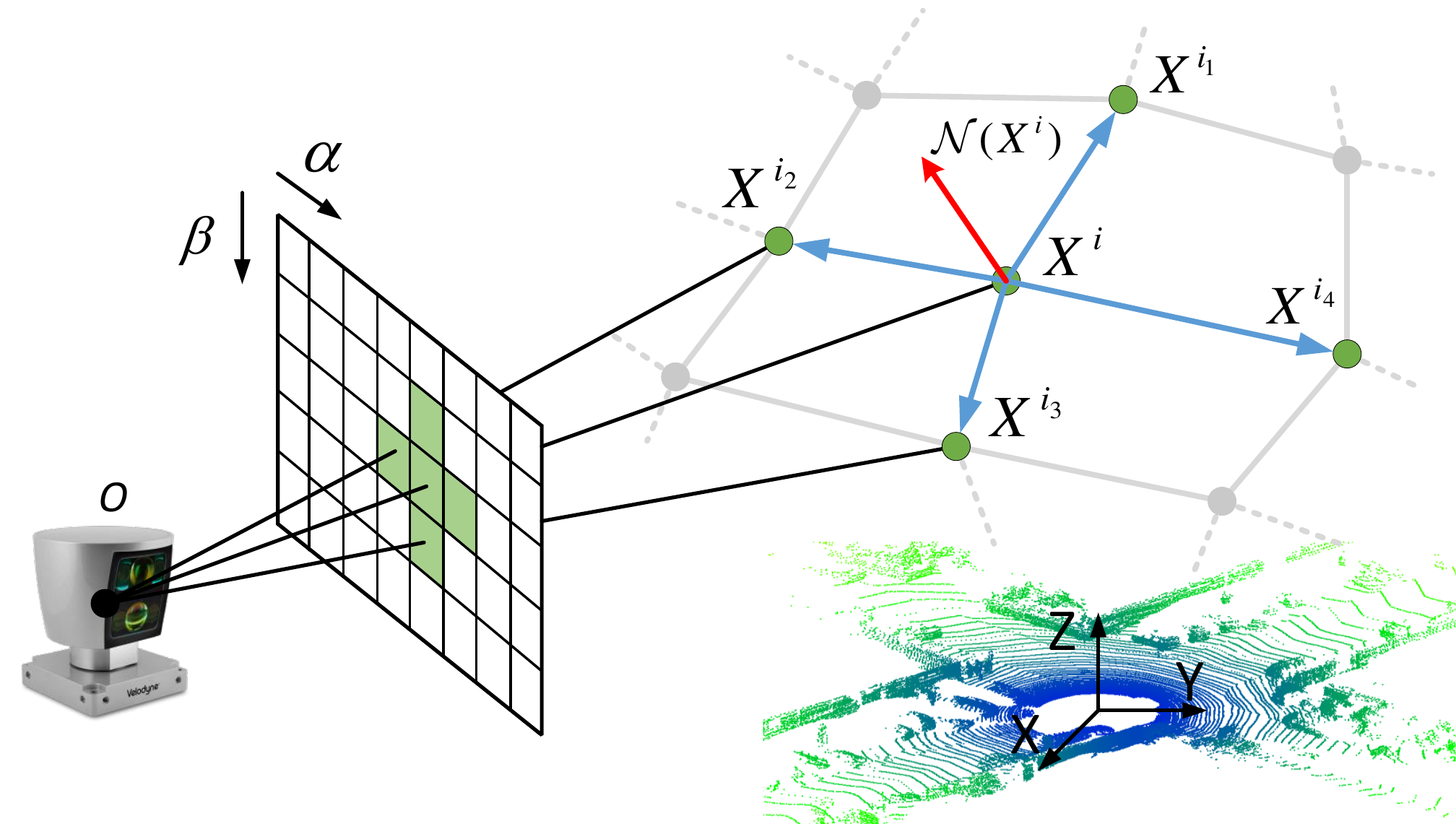}
	\end{center}
	\caption{Illustration of data encoding and normal estimation}
	\label{fig:project}
\end{figure}

As shown in Figure~\ref{fig:project}, the 3D lidar point clouds with ring structures are usually represented by Cartesian coordinates. 
Additional information includes intensity values. 
To convert the original sparse and irregular point clouds into a structured representation that can be fed into networks,
we encode the lidar data into point cloud matrices by a cylindrical projection~\cite{chen2017multi}. 
Given a 3D point $p = (x, y, z)$ in lidar coordinate system $(X, Y, Z)$,  the projection function is
\begin{equation}
\label{eq:projection}
\begin{array}{l}
	\alpha= \arctan(y/x) / \Delta \alpha   \\ 
	\beta= \arcsin(z/\sqrt{x^2+y^2+z^2}) / \Delta \beta 
\end{array}
\end{equation}
where $\alpha$ and $\beta$ are the indexes which set the points' positions in the matrix. 
$\Delta\alpha$ and $\Delta\beta$ are the average angular resolution between consecutive beam emitters 
in the horizontal and vertical directions, respectively.
The element at $(\alpha, \beta)$ of the point cloud matrix is filled with intensity value 
and range value $r=\sqrt{x^2+y^2+z^2}$ of the lidar point $p$. 
We keep the point closer to the lidar when multiple points are projected into a same position.
After applying this projection on the lidar data, we get a matrix of size $H\times W\times C$, and $C$ is the number of matrix channels. 
An example of the range channel of the matrix is shown in Figure~\ref{fig:mask}. 

\subsection{Geometric consistency constraint}    
\label{sec:normal}

\textbf{Normal estimation.}
As shown in Figure~\ref{fig:project}, given a 3D point $X^i$ and its $k$ neighbors $X^{i_j}$, $j = 1, 2, \ldots, k$ on the grid. 
The normal $\mathcal{N}(X^i)$ can be estimated by 
\begin{equation}
\label{eq:normal_1}
	\underset{\mathcal{N}(X^i)}{\arg\min}   \|[w_{i1}(X^{i_1}-X^i), \cdots, w_{ik}(X^{i_k}-X^i)]^T   \mathcal{N}(X^i)  \|_2
\end{equation}
where $(X^{i_k}-X^i)$ is a 3D vector, $w_{ik}$ is the weight of $X^{i_k}$ with respect to $X^i$, and $[\cdot]^T$ is a $k\times3$ vector. $\|\mathcal{N}(X^i)\|_2=1$. 
We set $w_{ik} = \exp(-0.2 |r(X^{i_k})-r(X^i)|)$ to put more weight on  points which have similar range value $r$ with $X^i$, and less weight otherwise. 
A commonly adopted strategy for solving Equation~(\ref{eq:normal_1}) is to perform a Principal Component Analysis (PCA), 
and convert the normal estimation to the computation of eigenvectors/eigenvalues of the covariance matrix created from $X^i$'s neighbors. 
In our work, this normal estimation needs to be embedded into the network.
The computation of covariance matrices and their eigenvectors makes the training and estimation inefficient. 
Hence, we simplify the normal estimation by computing the weighted cross products over $X^i$'s four neighbors. Then we smooth normal vectors using a moving average filter~\cite{moosmann2013interlacing}.
This can be formulated as 
\begin{equation}
\label{eq:normal}
	\mathcal{N}(X^i) = \sum_{X^{i_k}, X^{i_j}  \in  {\mathcal{P}}}   (w_{ik}(X^{i_k}-X^i) \times w_{ij}(X^{i_j}-X^i))
\end{equation}
where $\mathcal{P}$ is the set of neighboring points of $X^i$, sorted counterclockwisely, 
as $\{X^{i_1},X^{i_2}, X^{i_3}, X^{i_4}\}$ shown in Figure~\ref{fig:project}. 
The final normal vectors are normalized to 1.

Due to the temporal spatial geometric consistency of scan sequences, 
the points in a point cloud matrix should have the corresponding ones in another. 
Let $X_{t-1}^{\alpha\beta}$ and $X_t^{\alpha\beta}$ be the spatial corresponding point elements of the consecutive data matrices $S_{t-1}$ and $S_t$, respectively.
We can derive $\hat{X}_t^{\alpha\beta}$ from $X_{t-1}^{\alpha\beta}$ through
\begin{equation}
\label{eq:consistency}
	\hat{X}_t^{\alpha\beta}=P  T_t  P^{-1}  X_{t-1}^{\alpha\beta}
\end{equation}
where $T_t$ is the relative rigid pose transformation between the consecutive scans.
$P$ denotes the projection process and $P^{-1}$ is its inverse operation.
Therefore, $\hat{X}_t^{\alpha\beta}$ and $X_t^{\alpha\beta}$ are a pair of matching elements, 
and we can measure the similarity between the corresponding elements to verify the correctness of pose transformation. 
The noise inevitably exists in the coordinate and intensity values because of lidar measurements.
We compare the normal $\mathcal{N}(x)$ as it reflects smooth surface layouts and clear edge structures of the road environment (see Figure~\ref{fig:normal}). 
Thus, the constraint of pose transformation can be formulated as minimizing
\begin{equation}
\label{eq:consistency_loss}
	\mathcal{L}_n = \sum_{\alpha\beta}\|\mathcal{N}(\hat{X}_t^{\alpha\beta})  -  \mathcal{N}(X_t^{\alpha\beta})\|_1    \cdot    e^{|\nabla r(\hat{X}_t^{\alpha\beta})|}
\end{equation}
where $\nabla r(\hat{X}_t^{\alpha\beta})$ is a local range smooth measurement, 
and $\nabla$ is the differential operator with $\alpha$ and $\beta$. 
The item $e^{|\cdot|}$ allows the loss function to focus more on sharply changing areas in the scene.

\subsection{Lidar odometry regression}

To infer the 6-DoF relative pose between the scans $S_{t-1}$ and $S_t$, we construct a two-stream network. 
The input to the network is encoded point cloud matrices with point-wise normal vectors. 
As shown in Figure~\ref{fig:net}, LO-Net concatenates the features coming from two individual streams of feature extraction networks, 
then pass these concatenated features to the following four convolution units. 
The last three fully-connected layers output the 6-DoF pose between the input scans. 
The last two fully-connected layers are of dimensions 3 and 4, 
for regressing the translation \textit{x} and rotation quaternion \textit{q}, respectively. 
Finally, we get a 7D vector, which can be converted to a $4\times4$ transformation matrix. 

In order to reduce the number of model parameters and computation cost, 
we replace most of convolution layers of odometry regression network with fireConv~\cite{iandola2016squeezenet}, 
which has been used for object detection from lidar point clouds~\cite{wu2018squeezeseg}.
We follow \cite{iandola2016squeezenet} to construct our feature extraction layers. 
Since the width of intermediate features is much larger than its height, 
we only down-sample the width by using max-pooling during feature extraction. 
The detailed network architecture is illustrated in the supplementary material.

\medskip
\textbf{Learning position and orientation simultaneously.}
In our method, we choose the quaternion to represent the rotation as it is a continuous and smooth representation of rotation. 
The quaternions $q$ and $-q$ map to the same rotation, so we need to constrain them to a unit hemisphere.
We use $\mathcal{L}_x(S_{t-1};S_t)$ and $\mathcal{L}_q(S_{t-1};S_t)$ to demonstrate how to learn the relative translational 
and rotational components, respectively. 
\begin{equation}
\label{eq:loss_xq}
\begin{split}
	& \mathcal{L}_x(S_{t-1};S_t)=\|x_t-\hat{x}_t\|_l   \\
	& \mathcal{L}_q(S_{t-1};S_t)=\left\|q_t-\frac{\hat{q}_t}{\|\hat{q_t}\|}\right\|_l   \\
\end{split}
\end{equation}
where $x_t$ and $q_t$ are the ground truth relative translational and rotational components, 
$\hat{x}_t$ and $\hat{q}_t$ denote their predicted counterparts. $l$ refers to the distance norm in Euclidean space, 
and we use the $\ell_2$-norm in this work.  
Due to the difference in scale and units between the translational and rotational pose components,
previous works~\cite{kendall2015posenet, wang2017deepvo} gave a weight regularizer $\lambda$ to the rotational loss to jointly learn the 6-DoF pose. 
However, the hyper-parameter $\lambda$ need to be tuned when using new data from different scene.
To avoid this problem, we use two learnable parameters $s_x$ and $s_q$ to balance the scale between the translational and
rotational components in the loss term~\cite{kendall2017geometric}.
\begin{equation}
\begin{aligned}
\label{eq:loss_o}
	\mathcal{L}_o & = \mathcal{L}_x(S_{t-1};S_t)\exp(-s_x)+s_x \\
	& + \mathcal{L}_q(S_{t-1};S_t)\exp(-s_q)+s_q
\end{aligned}
\end{equation}
We use the initial values of $s_x=0.0$ and $s_q=-2.5$ for all scenes during the training.

\subsection{Mask prediction}

Given two consecutive scans $S_{t-1}$ and $S_t$ in a static rigid scene, 
we can get point matching relationships of the encoded data matrix pairs through transformation and cylindrical projection, 
as illustrated in Section~\ref{sec:normal}. 
Lidar point clouds are considered as the 3D model of the scene, and often contain dynamic objects such as cars and pedestrians in the road environment.
These factors may inhibit the learning pipeline of odometry regression.

Based on the encoder-decoder architecture (see Figure~\ref{fig:net}), 
we deploy a mask prediction network~\cite{zhou2017unsupervised, yang2017unsupervised} to learn the compensation for dynamic objects, 
and improve the effectiveness of the learned features and the robustness of the network.
The encoding layers share parameters with feature extraction layers of the odometry regression network,
and we train these two networks jointly and simultaneously.
The deconvolution layers are variants of fireDeconv~\cite{iandola2016squeezenet}, 
and adopt the skip-connection. 
The detailed network architecture is described in the supplementary material.

The predicted mask $\mathcal{M}(X_t^{\alpha\beta})\in[0,1]$ indicates the area where geometric consistency can be modeled or not, 
and implicitly ensures the reliability of the features learned in the pose regression network.
Therefore, the geometric consistency error as formulated in Equation~(\ref{eq:consistency_loss}) is weighted by
\begin{equation}
\label{eq:loss_n}
		\mathcal{L}_n = \sum_{\alpha\beta}  \mathcal{M}(X_t^{\alpha\beta})   \|\mathcal{N}(\hat{X}_t^{\alpha\beta})  -   \mathcal{N}(X_t^{\alpha\beta})\|_1    \cdot    e^{|\nabla r(\hat{X}_t^{\alpha\beta})|}   \ .
\end{equation}

Note that there is no ground truth label or supervision to train the mask prediction. 
The network can minimize the loss by setting all values of the predicted mask to be $0$.
To avoid this, we add a cross-entropy loss as a regularization term 
\begin{equation}
\label{eq:loss_r}
	\mathcal{L}_r=-\sum_{\alpha\beta}\log P(\mathcal{M}(X_t^{\alpha\beta})=1) .
\end{equation}

In summary, our final objective function to minimize for odometry regression is 
\begin{equation}
\label{eq:total_loss}
\mathcal{L} = \mathcal{L}_o + \lambda_n\mathcal{L}_n + \lambda_r\mathcal{L}_r
\end{equation}
where $\lambda_n$ and $\lambda_r$ are the weighting factors for geometric consistency loss and mask regularization, respectively.

\subsection{Mapping: scan-to-map refinement} 
\label{sec:mapping}

Consecutive scan-to-scan matches could introduce accumulative error, 
and also, may suffer when available common feature points between consecutive frames are limited. 
Hence, we maintain a global map reconstructed from the previously scans, 
then use the registration between this map and the current scan to refine the odometry. 
Unlike the traditional scan-to-map approaches~\cite{zhang2014loam, zhang2017low}, 
which directly match all the detected edge and plane points, 
we use the normal information (estimated by LO-Net) to select points from smooth area 
and use the mask (also an output of LO-Net) to exclude points from moving object.

\begin{figure}[t]
	\begin{center}
		\includegraphics[width=\linewidth]{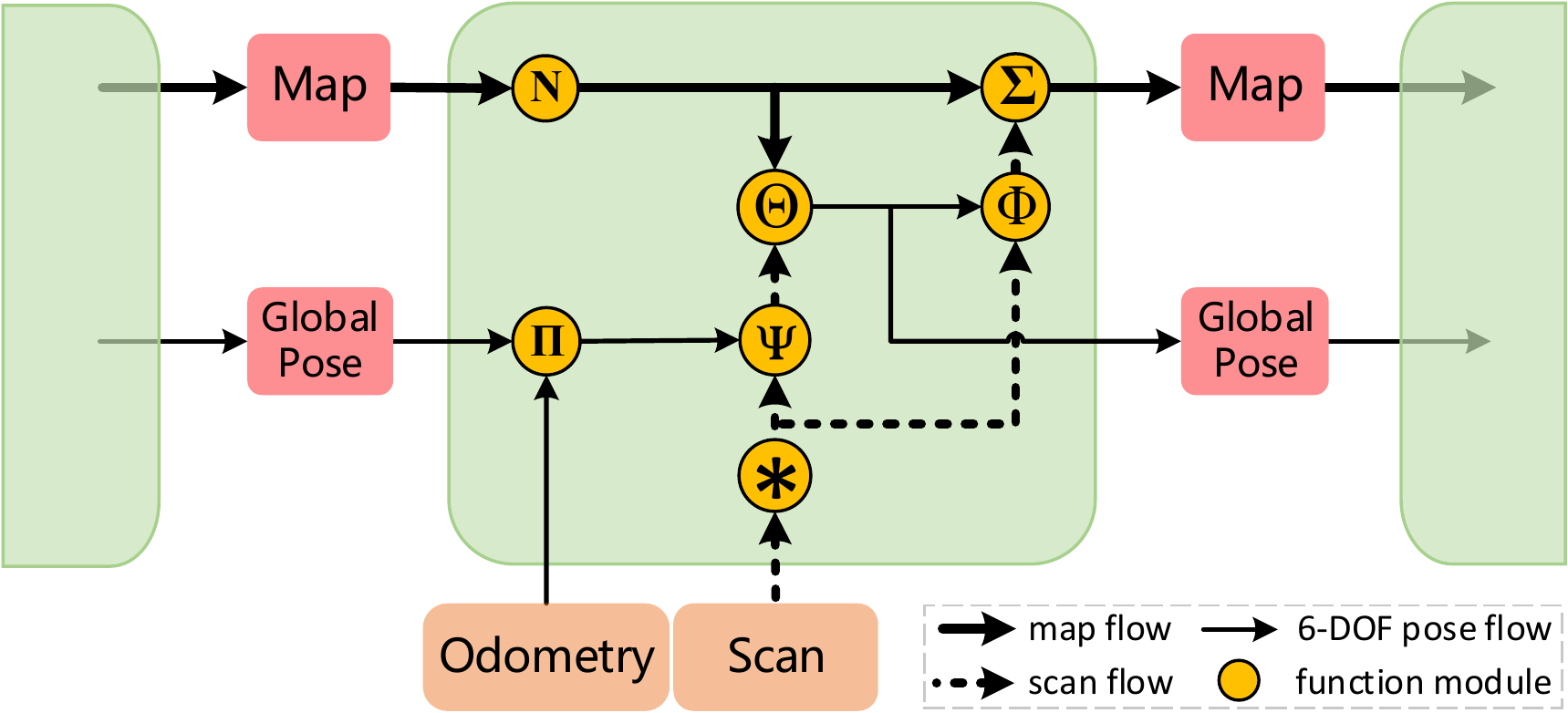}
	\end{center}
	\caption{The mapping module consecutively computes the low-drift motion of the lidar in the world coordinate system
		and builds a 3D map for the traversed environment using the lidar data. 
		The specific meanings of the function symbols are explained in the text.}
	\label{fig:mapping}
\end{figure}

At a test time $t$, let $[T_t, S_t]$ be the data from LO-Net.
$T_t$ is the odometry computed by LO-Net, which is used as the initial pose for mapping.
$S_t$ is a multi-channel data matrix containing the intensity, range, normal, and mask values of each point. 
The coordinates of each point can be calculated from its range value. 
The mapping takes the scan and the odometry as input, then matches the point cloud onto the global map. 
Figure~\ref{fig:mapping} shows the diagram of our mapping module. The details of diagram are as follows:

$\bm{\ast}$ :
Based on the normal channels of $S_t$,  we define a term $c$ to evaluate the smoothness of the local area. 
\begin{equation}
\label{eq:smoothness}
	c=\sum_{k=1}^3 (K\ast \mathcal{N}_k)^2 
\end{equation}
where $\mathcal{N}_k$ is the normal vector channel of $S_t$. The symbol $\ast$ denotes the convolution operation. 
$K$ is a $3\times5$ convolution kernel. For $K$, the value of the central element is -14, and the others are 1.
We compute $c$ for each point in $S_t$, and sort the values in increasing order. 
The first $n_c$ points in the list, except for marked points of the mask, are selected as planar points as they are in the smooth area.

$\bm{\Pi}$ :
Compute an initial estimate of the lidar pose relative to its first position: 
$\bm{{\rm M}}_{init} = \bm{{\rm M}}_{t-1}  \bm{{\rm M}}_{t-2} ^{-1} \bm{{\rm M}}_{t-1}$, 
where $\bm{{\rm M}}_t$ is the lidar transformation at time $t$.

$\bm{\Psi}$ :
Eliminate the motion distortion of lidar point cloud from $S_t$ by compensating for the ego-motion using a linear interpolation of $T_t$.  
Then, use $\bm{{\rm M}}_{init}$ to transform the corrected scan $S_t$ to the global coordinate system in which the map is located, 
and prepare for matching.

Suppose $\bm{p}_i=(p_{i_x}, p_{i_y}, p_{i_z}, 1)^T$ is a point in the scan $S_t$,
$\bm{m}_i=(m_{i_x}, m_{i_y}, m_{i_z}, 1)^T$ is the corresponding point in the map built by the previous scans, 
$\bm{n}_i = (n_{i_x}, n_{i_y}, n_{i_z}, 0)^T$ is the unit normal vector at $\bm{m}_i$. 
The goal of mapping is to find the optimal 3D rigid-body transformation
\begin{equation}
	\label{eq:point2plane}
	\bm{{\rm \hat{M}}}_{opt}= \underset{\bm{{\rm \hat{M}}}}{\arg\min}  \sum_i((\bm{{\rm \hat{M}}}  \cdot \bm{p}_i - \bm{m}_i)\cdot \bm{n}_i)^2    \ .
\end{equation} 

$\bm{\Theta}$ :
Iteratively register the scan onto the map by solving Equation~(\ref{eq:point2plane}) until a maximum number of iteration $n_{iter}$. 
Then, calculate the final transformation $\bm{{\rm M}}_t$ 
by accumulating the transformation during the iteration $\bm{{\rm \hat{M}}}_k$ and the initial pose $\bm{{\rm M}}_{init}$
\begin{equation}
\label{eq:trans_sum}
\bm{{\rm M}}_t = \prod_{k=1}^{n_{iter}}  \bm{{\rm \hat{M}}}_k  \bm{{\rm M}}_{init}   \ .
\end{equation}

$\bm{\Phi}$ :
Generate a new point cloud from the current scan $S_t$ by linear interpolation of vehicle motion between $\bm{{\rm M}}_{t-1}$ and $\bm{{\rm M}}_t$.

$\bm{\Sigma}$, \textbf{N}:
Add the new point cloud to the map.
Then, remove the oldest point cloud scans to only maintain $n_m$ transformed scans in the map.

This mapping-based refinement is performed iteratively along with the scan sequence. 
It can further improve the accuracy of odometry estimation, as shown in Section~\ref{sec:pose_result}.

\section{Experiments}

\textbf{Implementation details.}
The point cloud data we use is captured by the Velodyne HDL-64 3D lidar sensor(64 laser beams, 10Hz, collecting about 1.3 million points/second).
Therefore, during encoding the data matrix, we set $H=64$ and $W=1800$ by considering the sparseness of point clouds.
The width of input data matrix are resized to $1792$ by cropping both ends of the matrix.
During the training, the length of input sequence is set to be 3. 
We form the temporal pairs by choosing scan ${[S_{t-2}, S_{t-1}],  [S_{t-1}, S_{t}],  [S_{t-2}, S_{t}]}$. 
LO-Net predicts relative transformation between the pairs.
The whole framework is implemented with the popular Tensorflow library~\cite{abadi2016tensorflow}.
During the training, the mask prediction network is pre-trained using KITTI 3D object detection dataset, 
and all layers are trained simultaneously.
The initial learning rate is 0.001 and exponentially decays after every 10 epochs until 0.00001.  
The loss weights of Equation~(\ref{eq:total_loss}) are set to be $\lambda_n=0.15$ and $\lambda_r=0.05$, and the batch size is 8.
We choose the Adam~\cite{kingma2014adam} solver with default parameters for optimization. 
The network is trained on an NVIDIA 1080 Ti GPU.
For the mapping, we set the number of points in each scan $n_c=0.01HW$, 
the number of scans in the map $n_m = 100$, and the number of iterations $n_{iter}=15$.

\begin{figure}[t]
	\begin{center}
		\includegraphics[width=\linewidth]{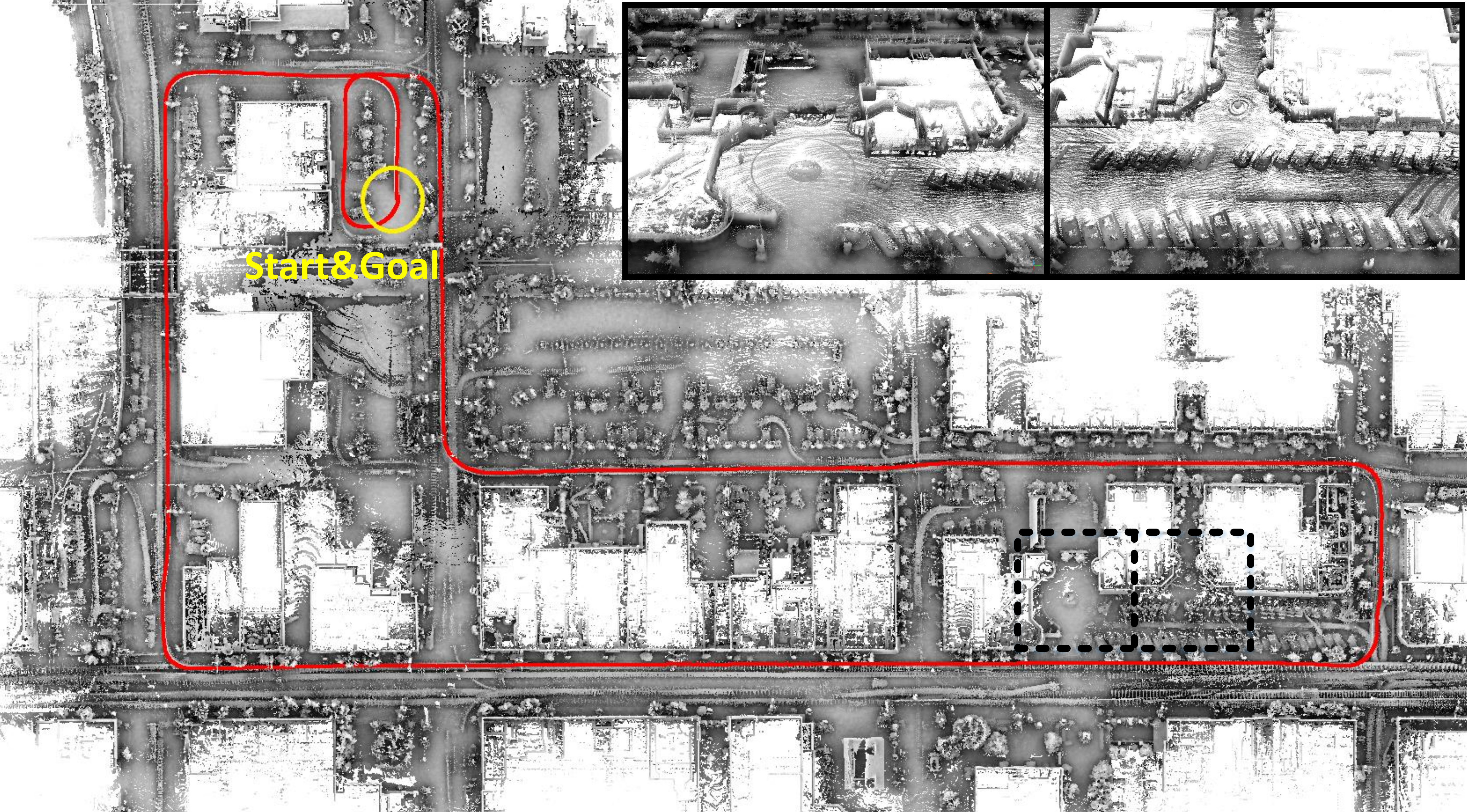}
	\end{center}
	\caption{Trajectory (in red) and the built map of LO-Net+Mapping on Ford dataset 1. 
		Our reconstructed trajectory (without enforcing loop closure) has small drift and forms closed loops accurately  (yellow circle). }
	\label{fig:Ford-1}
\end{figure}

\subsection{Datasets}

\textbf{KITTI.}
The KITTI odometry dataset~\cite{geiger2013vision, geiger2012we} consists of 22 independent sequences with 
stereo gray-scale and color camera images, point clouds captured by a Velodyne lidar sensor, and calibration files. 
Sequences 00-10 (23201 scans) are provided with ground truth pose obtained from the IMU/GPS readings.
For sequences 11-21 (20351 scans), there is no ground truth available, and are provided for benchmarking purposes.
The dataset is captured during driving in a variety of road environments with dynamic objects and vegetation, 
such as highways, country roads and urban areas. The driving speed is up to 90km/h.

\medskip
\textbf{FORD.}
The Ford Campus Vision and Lidar Data Set~\cite{pandey2011ford} consists time-synchronized 2D panoramic image, 
3D lidar point clouds and IMU/GPS data. 
Like KITTI, the lidar dataset is captured using horizontally scanning 3D lidar mounted on the top of a vehicle. 
The dataset contains two loop closure sequences collected in different urban environments, 
and there are more moving vehicles than the KITTI dataset.

\begin{figure}[t]
	\begin{center}
		\includegraphics[width=\linewidth]{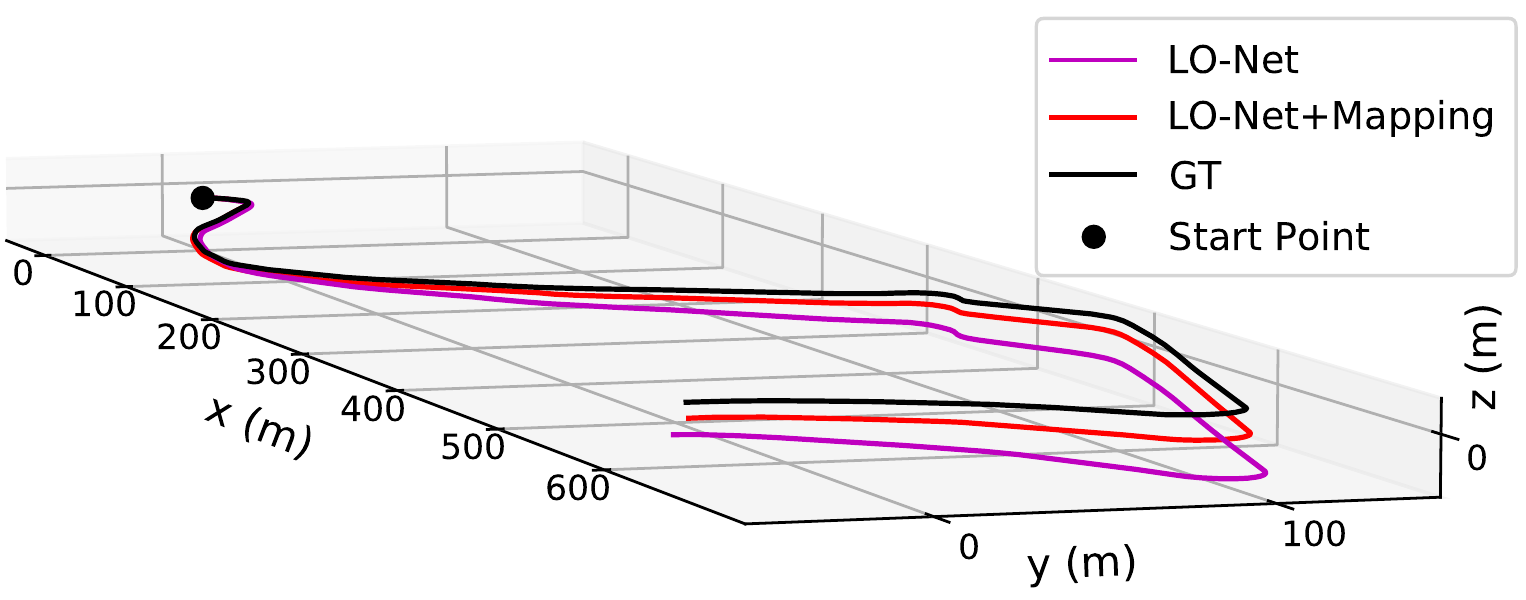}
	\end{center}
	\caption{3D trajectory plots of our method for KITTI Seq. 10. 
		The mapping module effectively reduces the vertical drift generated in LO-Net.}
	\label{fig:seq10path}
\end{figure}

\subsection{Odometry results}   
\label{sec:pose_result}

\begin{table*}[t]
	\centering	
	\caption{Odometry results on KITTI and Ford datasets. 
		Our network is trained on KITTI sequences and then tested on the two datasets.}
	\label{tab:pose}
	\footnotesize
	\resizebox{\linewidth}{!}{
	\begin{threeparttable}
		\begin{tabular}{ccccccccccccc|cccc}
			\toprule
			\multirow{2}{*}{Seq.} &\multicolumn{2}{c}{ICP-po2po} &\multicolumn{2}{c}{ICP-po2pl} &\multicolumn{2}{c}{GICP~\cite{segal2009generalized}} &\multicolumn{2}{c}{CLS~\cite{velas2016collar}} &\multicolumn{2}{c}{LOAM~\cite{zhang2017low}\tnote{1}} &\multicolumn{2}{c}{Velas \etal~\cite{velas2018cnn}\tnote{2}} &\multicolumn{2}{c}{LO-Net} &\multicolumn{2}{c}{LO-Net+Mapping}\\  \cline{2-17}                  
			&$t_{rel}$&$r_{rel}$&$t_{rel}$&$r_{rel}$&$t_{rel}$&$r_{rel}$&$t_{rel}$&$r_{rel}$&$t_{rel}$&$r_{rel}$&$t_{rel}$&$r_{rel}$&$t_{rel}$&$r_{rel}$&$t_{rel}$&$r_{rel}$\\
			\midrule
			00\tnote{\dag}   &6.88   &2.99    &3.80  &1.73    &1.29  &0.64    &2.11  &0.95    &1.10\ (\textbf{0.78})  &0.53     &3.02  &NA    &1.47  &0.72    &\textbf{0.78}  &\textbf{0.42}  \\
			01\tnote{\dag}   &11.21  &2.58    &13.53 &2.58    &4.39  &0.91    &4.22  &1.05    &2.79\ (1.43)  &0.55     &4.44  &NA    &\textbf{1.36}  &0.47    &1.42  &\textbf{0.40}  \\
			02\tnote{\dag}   &8.21   &3.39    &9.00  &2.74    &2.53  &0.77    &2.29  &0.86    &1.54\ (\textbf{0.92})  &0.55     &3.42  &NA    &1.52  &0.71    &1.01  &\textbf{0.45}  \\
			03\tnote{\dag}   &11.07  &5.05    &2.72  &1.63    &1.68  &1.08    &1.63  &1.09    &1.13\ (0.86)  &0.65     &4.94  &NA    &1.03  &0.66    &\textbf{0.73}  &\textbf{0.59}  \\
			04\tnote{\dag}   &6.64   &4.02    &2.96  &2.58    &3.76  &1.07    &1.59  &0.71    &1.45\ (0.71)  &\textbf{0.50}     &1.77  &NA    &\textbf{0.51}  &0.65    &0.56  &0.54  \\
			05\tnote{\dag}   &3.97   &1.93    &2.29  &1.08    &1.02  &0.54    &1.98  &0.92    &0.75\ (\textbf{0.57})  &0.38     &2.35  &NA    &1.04  &0.69    &0.62  &\textbf{0.35}  \\
			06\tnote{\dag}      &1.95   &1.59    &1.77  &1.00    &0.92  &0.46    &0.92  &0.46    &0.72\ (0.65)  &0.39     &1.88  &NA    &0.71  &0.50    &\textbf{0.55}  &\textbf{0.33}  \\
			07\tnote{*}      &5.17   &3.35    &1.55  &1.42    &0.64  &0.45    &1.04  &0.73    &0.69\ (0.63)  &0.50     &1.77  &NA    &1.70  &0.89    &\textbf{0.56}  &\textbf{0.45}  \\
			08\tnote{*}      &10.04  &4.93    &4.42  &2.14    &1.58  &0.75    &2.14  &1.05    &1.18\ (1.12)  &0.44     &2.89  &NA    &2.12  &0.77    &\textbf{1.08}  &\textbf{0.43}  \\
			09\tnote{*}      &6.93   &2.89    &3.95  &1.71    &1.97  &0.77    &1.95  &0.92    &1.20\ (\textbf{0.77})  &0.48     &4.94  &NA    &1.37  &0.58    &\textbf{0.77}  &\textbf{0.38}  \\
			10\tnote{*}      &8.91   &4.74    &6.13  &2.60    &1.31  &0.62    &3.46  &1.28    &1.51\ (\textbf{0.79})  &0.57     &3.27  &NA    &1.80  &0.93     &0.92  &\textbf{0.41}  \\ 
			\hline
			mean\tnote{\dag} &7.13&3.08  &5.15&1.91	&2.23&0.78	&2.11&0.86	&1.35\ (0.85)&0.51 &3.12&NA&1.09&0.63	&\textbf{0.81}&\textbf{0.44} \\
			mean\tnote{*}        &7.76&3.98  &4.01&1.97	&1.38&0.65	&2.15&1.00	&1.15\ (\textbf{0.83})&0.50 &3.22&NA&1.75&0.79	&\textbf{0.83}&\textbf{0.42} \\
			\hline \hline
			Ford-1      &8.20&2.64       &3.35&1.65   &3.07&1.17     &10.54&3.90   &1.68&0.54   &NA&NA    &2.27&0.62	 &\textbf{1.10}&\textbf{0.50}  \\
			Ford-2      &16.23&2.84 	&5.68&1.96   &5.11&1.47     &14.78&4.60	  &1.78&0.49   &NA&NA    &2.18&0.59 	&\textbf{1.29}&\textbf{0.44}  \\
			\bottomrule
		\end{tabular}
		\begin{tablenotes}
			\item $^1$:  The results on KITTI dataset outside the brackets are obtained by running the code, and those in the brackets are taken from~\cite{zhang2017low}.
			\item $^2$:  The results on KITTI dataset are taken from~\cite{velas2018cnn}, and the results on Ford dataset are not available.
			\item $^{\dag}$: The sequences of KITTI dataset that are used to train LO-Net.
			\item $^*$: The sequences of KITTI dataset that are not used to train LO-Net.
			\item $t_{rel}$: Average translational RMSE $(\%)$ on length of 100m-800m.
			\item $r_{rel}$: Average rotational RMSE ($^\circ/100$m) on length of 100m-800m.
		\end{tablenotes}
	\end{threeparttable}}
\end{table*}

\textbf{Baselines.}
We compare our approach with several classic lidar odometry estimation methods: 
ICP-point2point (ICP-po2po), ICP-point2plane (ICP-po2pl), GICP~\cite{segal2009generalized}, 
CLS~\cite{velas2016collar}, LOAM~\cite{zhang2017low} and Velas \etal~\cite{velas2018cnn}. 
The first two ICP methods are implemented using the Point Cloud Library~\cite{rusu20113d}.
As far as we know, \cite{velas2018cnn} is the only deep learning based lidar odometry method that has comparable results, 
but it has no codes publicly available.
We obtain the results of other evaluated methods using the publicly available code, 
and some of the results are even better than those in the originally published papers.
For LOAM algorithm, it achieves the best results among lidar-based methods in the KITTI odometry evaluation benchmark~\cite{geiger2012we}.
In order to enable the map-based optimization of LOAM to run for every input scan and determine the full potential of the algorithm, 
we make some modifications and parameter adjustments in the originally published code.
Loop closure detection is not implemented for all methods since we aim to test the limits of accurate odometry estimation.

\begin{figure}[t]
	\begin{center}
		\includegraphics[width=\linewidth]{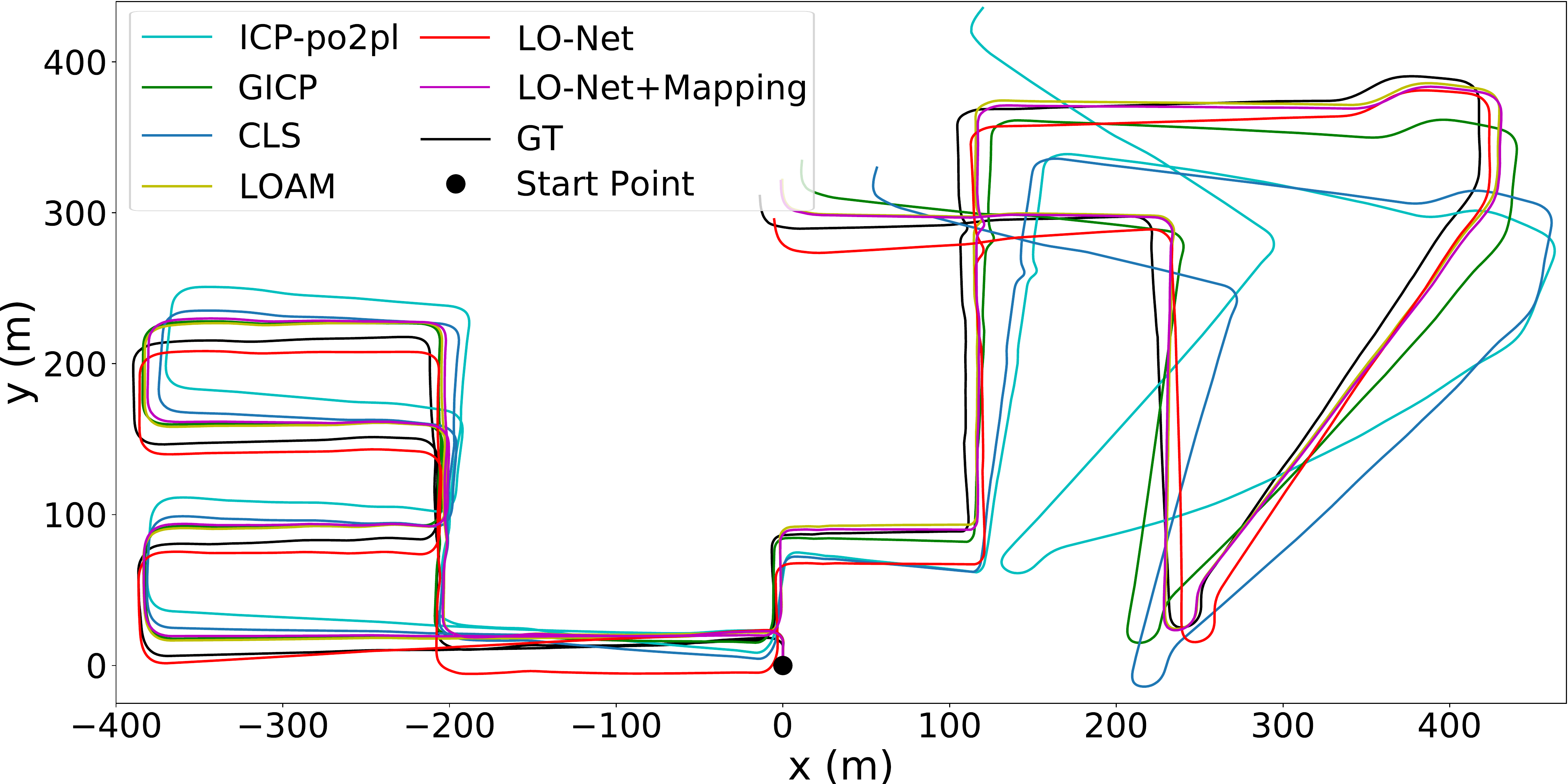}
	\end{center}
	\caption{Trajectory plots of KITTI Seq. 08 with ground truth. 
		Our LO-Net+Mapping produces most accurate trajectory.}
	\label{fig:seq08path}
\end{figure}

\begin{figure}[t]
	\begin{center}
		\includegraphics[width=\linewidth]{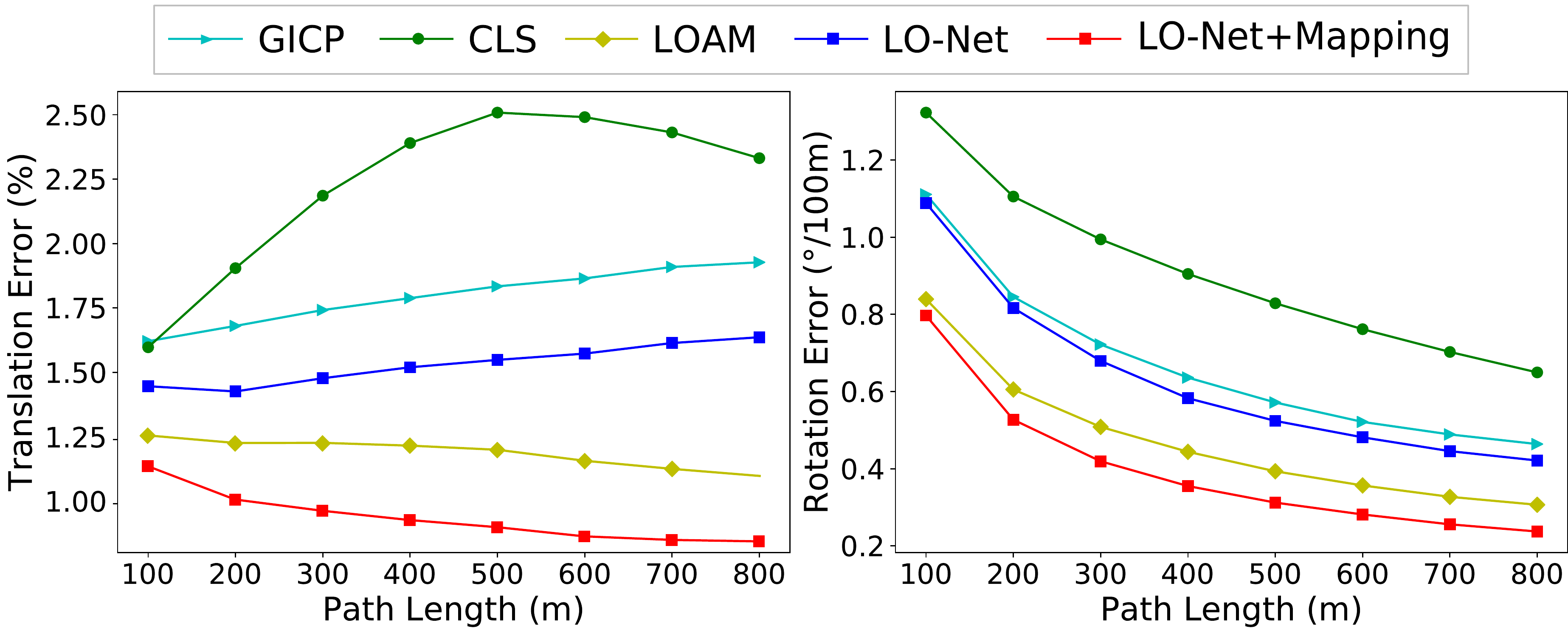}
	\end{center}
	\caption{Evaluation results on the KITTI Seq. 00-10. 
		We show the average errors of translation and rotation with respect to path length intervals.  
		Our LO-Net+Mapping achieves the best performance among all the evaluated methods.}
	\label{fig:segError}
\end{figure}

We firstly conduct the training and testing experiments on the KITTI dataset. 
Then, based on the model trained only on the KITTI dataset, 
we directly test the model on the Ford dataset without any further training or fine-tuning.
We use the KITTI odometry evaluation metrics~\cite{geiger2012we} to quantitatively analyze the accuracy of odometry estimation.
Table~\ref{tab:pose} shows the evaluation results of the methods on KITTI and Ford datasets. 
It can be seen that the results of LO-Net+Mapping are slightly better than LOAM and clearly superior to the others.
Although there are differences between the two datasets, 
such as different lidar calibration parameters and different systems for obtaining ground truth,
our approach still achieves the best average performance among evaluated methods. 
Figure~\ref{fig:seq10path} shows the 3D trajectory plots of our method at different stages.
Some trajectories produced by different methods are shown in Figure~\ref{fig:Ford-1} and ~\ref{fig:seq08path}.
Figure~\ref{fig:segError} shows the average evaluation errors on KITTI Seq. 00-10.
More estimated trajectories for KITTI and Ford datasets are shown in our supplementary material.

\medskip
\textbf{Ablation study.}
In order to investigate the importance of different loss components proposed in Section~\ref{sec:method}, 
we perform an ablation study on KITTI dataset by training and testing the LO-Net using different combinations of the losses.
As shown in Table~\ref{tb:loss}, the network achieves the best average performance 
when it is trained with the full loss.

\begin{table}[t]
	\caption{Comparison of different combinations of the losses. 
		The mean values of translational and rotational RMSE on KITTI training and testing sequences are computed as in Table~\ref{tab:pose}.
		$\mathcal{L}_n'$ indicates that the geometric consistency loss is not weighted by the mask.}
	\label{tb:loss}
	\footnotesize
	\centering
	\resizebox{\linewidth}{!}{
		\begin{tabular}{ccccccccccc}
			\toprule
			\multirow{2}{*}{Seq.} & &\multicolumn{2}{c}{$\mathcal{L}_o$} & &\multicolumn{2}{c}{$\mathcal{L}_o$, $\mathcal{L}_n'$}  & &\multicolumn{2}{c}{$\mathcal{L}_o$, $\mathcal{L}_n$, $\mathcal{L}_r$}  &    \\  \cline{2-11} 
			&&$t_{rel}$&$r_{rel}$  & &$t_{rel}$&$r_{rel}$   & &$t_{rel}$&$r_{rel}$  &   \\
			\midrule
			mean$^{\dag}$  &     &1.46   &1.01    &     &1.18 &0.70     &     &\textbf{1.09} &\textbf{0.63}   & \\
			mean$^*$           &     &2.03   &1.50    &     &1.80 &0.82     &     &\textbf{1.75} &\textbf{0.79}   & \\
			\bottomrule
	\end{tabular}}
\end{table}

\subsection{Normal results}
\begin{figure}[t]
	\begin{center}
		\includegraphics[width=\linewidth]{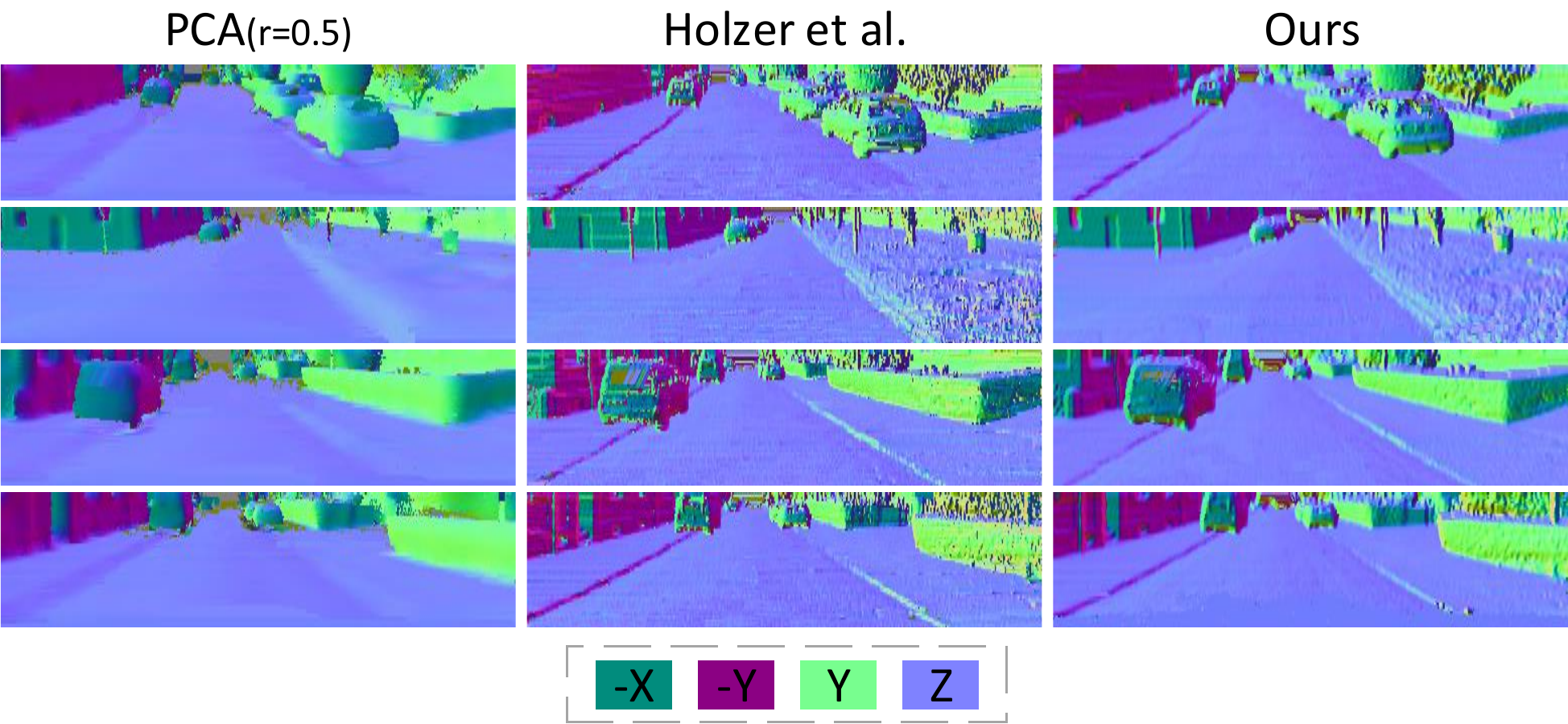}
	\end{center}
	\caption{Visual comparison of normal results on KITTI dataset. 
		Different colors indicate different normal directions. 
		Our results show smooth surface layouts and clear edge structures. 
		The images are cropped and reshaped for better visualization.}
	\label{fig:normal}
\end{figure}

Since KITTI and Ford datasets do not provide a benchmark for normal evaluation, 
we compare the normal estimation with that computed from the methods of PCA and Holzer~\cite{holzer2012adaptive}.
The PCA estimates the surface normal at a point by fitting a least-square plane from its surrounding neighboring points.
In our experiment, we choose the radius $r=0.5$m and $r=1.0$m as the scale factor to determine the set of nearest neighbors of a point.
As shown in Figure~\ref{fig:normal}, our estimated normals can extract smooth scene layouts and clear edge structures.

For quantitative comparison purpose, the normals computed from PCA with $r=0.5$m are interpolated and used as the ground truth. 
Then a point-wise cosine distance is employed to compute the error between the predicted normal and the ground truth. 
\begin{equation}
\label{eq:normalEval}
	e_i = \arccos (\bm{n}_i \cdot \hat{\bm{n}}_i), \quad \hat{\bm{n}}_i \in \mathcal{N}
\end{equation}
where the angle $e_i$ is the normal error at point $p_i$, $\bm{n}_i$ and $\hat{\bm{n}}_i$ are ground truth and predicted normal vector of point $p_i$, respectively. 
$\hat{\bm{n}}_i \in \mathcal{N}$ denotes the point $p_i$ is a valid point with ground truth normal. 
The normal evaluations performed on KITTI dataset are shown in Table~\ref{tb:normal}, 
our approach outperforms the others under most metrics. 
The metrics include the mean and median values of $e_i$, 
and the percent-good-normals whose angle fall within the given thresholds~\cite{fouhey2013data, yang2018lego}. 
\textquotedblleft GT median\textquotedblright \ which denotes that 
we set a normal direction for all points with the median value of the ground truth, is employed as a baseline. 
The evaluation demonstrates that our estimated normal can serve as a reliable property of road scene for the geometric consistency constraint. 

\begin{table}[t]
	\caption{Normal performances of our method and the baseline methods on KITTI dataset. }
	\label{tb:normal}
	\footnotesize
	\centering
	\resizebox{\linewidth}{!}{
	\begin{tabular}{cccccc}
		\toprule
		\multirow{2}{*}{Method}&\multicolumn{2}{c}{(Lower Better)} &\multicolumn{3}{c}{(Higher Better)}  \\
		                                & Mean     &  Median     & $<11.25^{\circ} $    & $<22.5^{\circ} $   & $<30^{\circ} $ \\
		\midrule
		GT median                &23.38	  &5.78   &0.632   &0.654   &0.663 \\
		PCA(r=1.0)             &14.38	  &11.55   &0.470   &\textbf{0.895}   &\textbf{0.946} \\
		Holzer \etal~\cite{holzer2012adaptive}        &13.18	 &5.19   &0.696   &0.820   &0.863 \\
		Ours                        &\textbf{10.35}	   &\textbf{3.29}   &\textbf{0.769}   &0.865   &0.897  \\
		\bottomrule
	\end{tabular}}
\end{table}

\subsection{ Mask visualization }
\begin{figure}[t]
	\begin{center}
		\includegraphics[width=\linewidth]{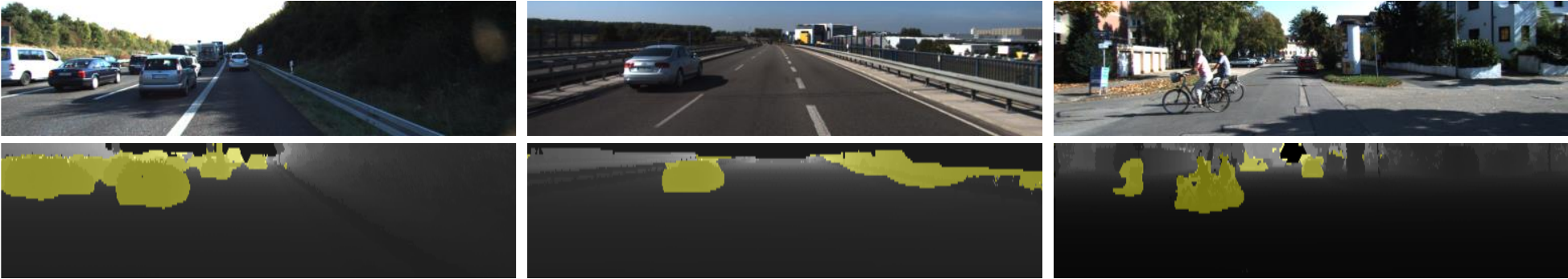}
	\end{center}
	\caption{Sample visualizations of masks on range channel of the data matrix and its corresponding RGB images. 
		The yellow pixels indicate the uncertain points in the surrounding environment for odometry estimation, 
		such as points of moving cars, cyclists and others. 
		The images are cropped and reshaped for better visualization.}
	\label{fig:mask}
\end{figure}

Examples of the mask predicted by our network are visualized in Figure~\ref{fig:mask}. 
The highlighted areas suggest that LO-Net has learned to identify dynamic objects and tend to mask vegetation as unexplainable, 
and indicate that the network will pay less attention to these areas in the odometry regression. 
Dynamic objects and the relationships between scan sequences are important for odometry estimation problems. 
They are difficult to explicitly model but implicitly learned by our network.

\subsection{Runtime}
The lidar scanning point cloud are captured one by one over time, 
and processing these data in time is critical for robotic applications. 
Note that unlike image-based computer vision applications, 
the commonly used lidar sensors, such as Velodyne HDL-64 used in the KITTI and Ford dataset, 
rotate at a rate of 10Hz, that is 0.1s per-scan.
Therefore, the real-time performance here means that the processing time of each scanning data is less than 0.1s.
An NVIDIA 1080 Ti GPU and an Intel Core i7 3.4GHz 4-core CPU are chose as our test platform.
In test-time, the data batch-size of LO-Net is set to 1.
Table~\ref{tb:time} shows the average running times on Seq. 00 of KITTI dataset.
The average processing time of our framework is about 80.1ms per-scan in total.
Reasonably, most of our runtime is spent on the mapping procedure.
Compared with most of traditional lidar odometry estimation methods, 
including the methods evaluated in Section~\ref{sec:pose_result}, 
our map-based optimization is lightning fast since we consume a new representation of input data. 
Our approach enables real-time performance through a straight-forward pipeline on a platform with GPU.
For lower performance platforms, we can also speed up the processing through the parallelism of LO-Net and the mapping. 
Currently, some parts of our framework run on CPU, and we can implement them on GPU to further increase the speed.

\begin{table}[t]
\caption{Average runtime on KITTI Seq. 00}
\label{tb:time}
\small
\centering
\resizebox{\linewidth}{!}{
\begin{tabular}{cccc}
	\toprule
	Data preparation    & Inference                 &  Mapping                & total     \\
	\midrule
	8.5ms on CPU         & 10.2ms on GPU       &  61.4ms on CPU     & 80.1ms  \\
	\bottomrule
\end{tabular}}
\end{table}

\section{Conclusions}
We present a novel learning framework LO-Net to perform lidar odometry estimation. 
An efficient mapping module is coupled into the estimation framework to further improve the performance. 
Experiments on public benchmarks demonstrate the effectiveness of our framework over existing approaches. 

There are still some challenges that need to be addressed:  
1) The point clouds are encoded into data matrices to feed into the network. 
Direct processing of 3D point clouds could be more practical for 3D visual tasks. 
2) Our current network is trained with ground truth data. 
This limits the application scenarios of the network. 
In our future work, we will investigate in more detail the geometry feature representation learned by the network. 
We also plan to incorporate recurrent units into this network to build temporal-related features. 
This may lead to an end-to-end framework without the need of costly collections of ground truth data. 

\section*{Acknowledgment}
This work is supported by National Natural Science Foundation of China (No. U1605254, 61728206), and the National Science Foundation of USA under Grants EAR-1760582.

{\small
\bibliographystyle{ieee_fullname}
\bibliography{refbib}
}


\clearpage
\appendix
\section*{Supplementary Material}
\medskip




\section{Network parameters}

\begin{figure}[th]
	\centering
	\includegraphics[width=0.85\linewidth]{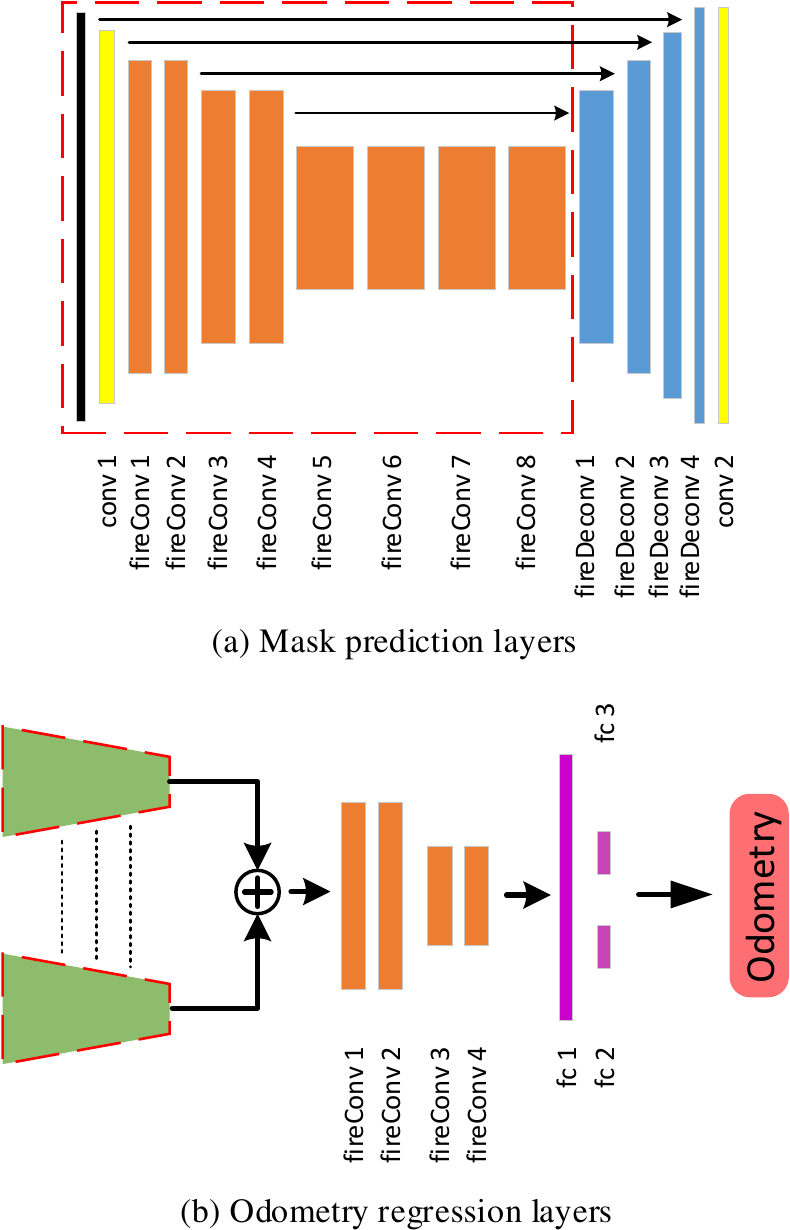}
	\caption{The mask prediction layers and odometry regression layers of our LO-Net.}
	\label{fig:net_part}
\end{figure}

\begin{figure}[th]
	\centering
	\includegraphics[width=0.6\linewidth]{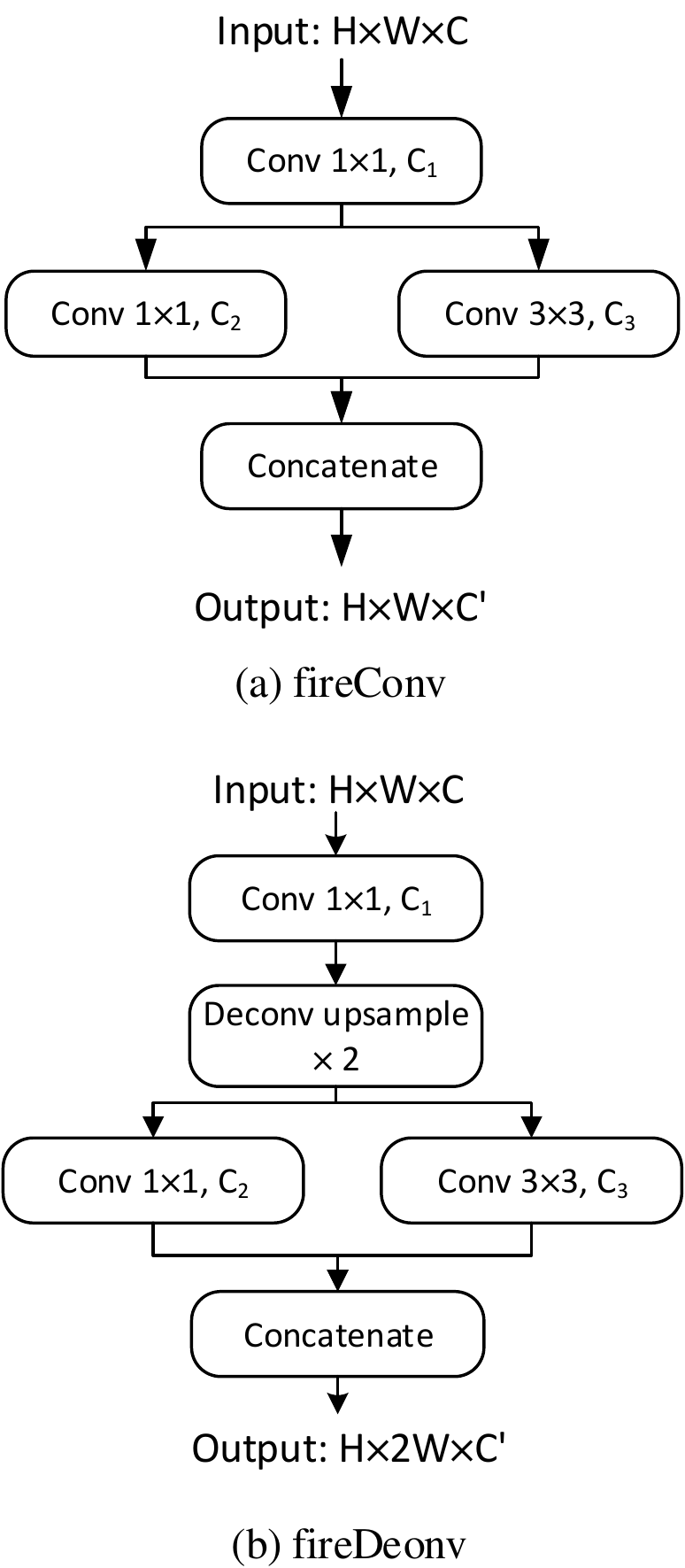}
	\caption{The structure of fireConv and fireDeconv from SqueezeNet.}
	\label{fig:fireModule}
\end{figure}

Figure~\ref{fig:net_part} show the mask prediction layers and odometry regression layers of our LO-Net. 
In order to reduce the number of model parameters and computation cost, 
and make it is possible to run the network on a low performance platform, such as mobile robot or backpack system. 
We replace most of convolutional layers of the network with fireConv, and deconvolutional layers with fireDeconv. 
The fireConv and fireDeconv module of SqueezeNet~\cite{iandola2016squeezenet} can construct a light-weight network that can achieve similar performance as AlexNet~\cite{krizhevsky2012imagenet}. 
Structures of the two modules are shown in Figure~\ref{fig:fireModule}. 
The first $1\times1$ convolution compresses the input tensor. 
The second $1\times1$ convolution and the $3\times3$ convolution let the network to learn more feature representations from different kernel sizes.  

Since the width of intermediate features is much larger than its height, 
we only down-sample the width by maxpooling during feature extraction. 
The reweighing layer~\cite{wang2018pointseg} is adopted to learn a more robust feature representation. 
For mask prediction, the fireDeconv is used to up-sample the feature maps and get the original scale resolution point-wise mask prediction. 
To infer the 6-DoF relative pose between the two input scans, 
we concatenate output features from the enlargement layer~\cite{wang2018pointseg} of mask prediction layers.
The last two fully-connected layers output the translation $x$ and rotation quaternion $q$, respectively.

The network parameters are shown in Table~\ref{tab:net1} and \ref{tab:net2}.
All convolutional and deconvolutional layers are followed by Rectified Linear Unit (ReLU)~\cite{glorot2011deep} except for the last output layers where nonlinear activation function is applied. 
We experimented with batch normalization performed on convolutional layers, data normalization and rescaling,  
however they did not yield significant performance gains, rather in some cases they negatively affected the odometry accuracy.

\newcommand{\tabincell}[2]{\begin{tabular}{@{}#1@{}}#2\end{tabular}}

\begin{table}[t]
	\centering
	\caption{Network parameters of mask prediction layers.
		For fireConv and fireDeconv layers, the filter size is represented as $C_1-C_2-C_3$ as shown in Figure~\ref{fig:fireModule}.}
	\label{tab:net1}
	\resizebox{\linewidth}{!}{
	\begin{tabular}{cccccc}
		\toprule
		Layer  &&              Filter size &&                   Kernel size / Stride &  \\ 
		\midrule
		conv 1&&	                                   64&&	                              3 / 1$\times$2& \\
		maxpooling&&	                            &&                               	3 / 1$\times$2& \\
		fireConv 1&&                    16-64-64&&                                  &\\	
		fireConv 2&&                    16-64-64&&                                   &\\	
		\tabincell{c}{maxpooling + \\ reweighing 1}  &&	                &&                 3 / 1$\times$2& \\
		fireConv 3&&                32-128-128&&                                   	& \\
		fireConv 4&&              32-128-128&&                                     &	 \\	
		\tabincell{c}{maxpooling + \\ reweighing 2}  &&                	&&                 3 / 1$\times$2& \\
		fireConv 5&&               48-192-192&&                                             & \\	
		fireConv 6&&              48-192-192&&              				            & \\	
		fireConv 7&&	          64-256-256&&             		                         &  \\	
		fireConv 8&&	          64-256-256&&             		                            &  \\	
		\tabincell{c}{enlargement + \\ reweighing 3}  &&                &&                             & \\
		fireDeconv 1&&         64-128-128&&               		                    & \\	
		fireDeconv 2&&    	       64-64-64&&                      				           &  \\	
		fireDeconv 3&&	           16-32-32&&                      			             &  \\	
		fireDeconv 4&&	           16-32-32&&                      			             &  \\	
		dropout (0.5)      &&                           &&                                                       & \\
		conv 2       &&   	                     2 &&	                				    3 / 1$\times$1& \\
		\bottomrule
	\end{tabular}}
\end{table}

\begin{table}[t]
	\centering
	\caption{Network parameters of odometry regression layers.}
	\label{tab:net2}
	\resizebox{\linewidth}{!}{
	\begin{tabular}{cccccc}
		\toprule
		Layer  &&                    Filter size &&                   Kernel size / Stride &  \\ 
		\midrule
		fireConv 1&&                    64-256-256&&                                  & \\	
		fireConv 2&&                    64-256-256&&                                   & \\	
		\tabincell{c}{maxpooling + \\ reweighing 1}  &&	              	                &&                         	3 / 2$\times$2& \\
		fireConv 3&&                   80-384-384&&                                   	& \\
		fireConv 4&&                   80-384-384&&                                     & \\	
		maxpooling                            &&                              	&&                            	3 / 2$\times$2& \\
		fc 1&&                             512&&                                                      & \\	
		dropout (0.5) &&                           &&                                                       & \\
		fc 2&&                                 3&&              				                         & \\	
		fc 3&&	                               4&&             		                                     &  \\	
		\bottomrule
	\end{tabular}}
\end{table}

\section{More odometry results}

\begin{figure*}[!h]
	\centering
	\includegraphics[width=0.9\linewidth]{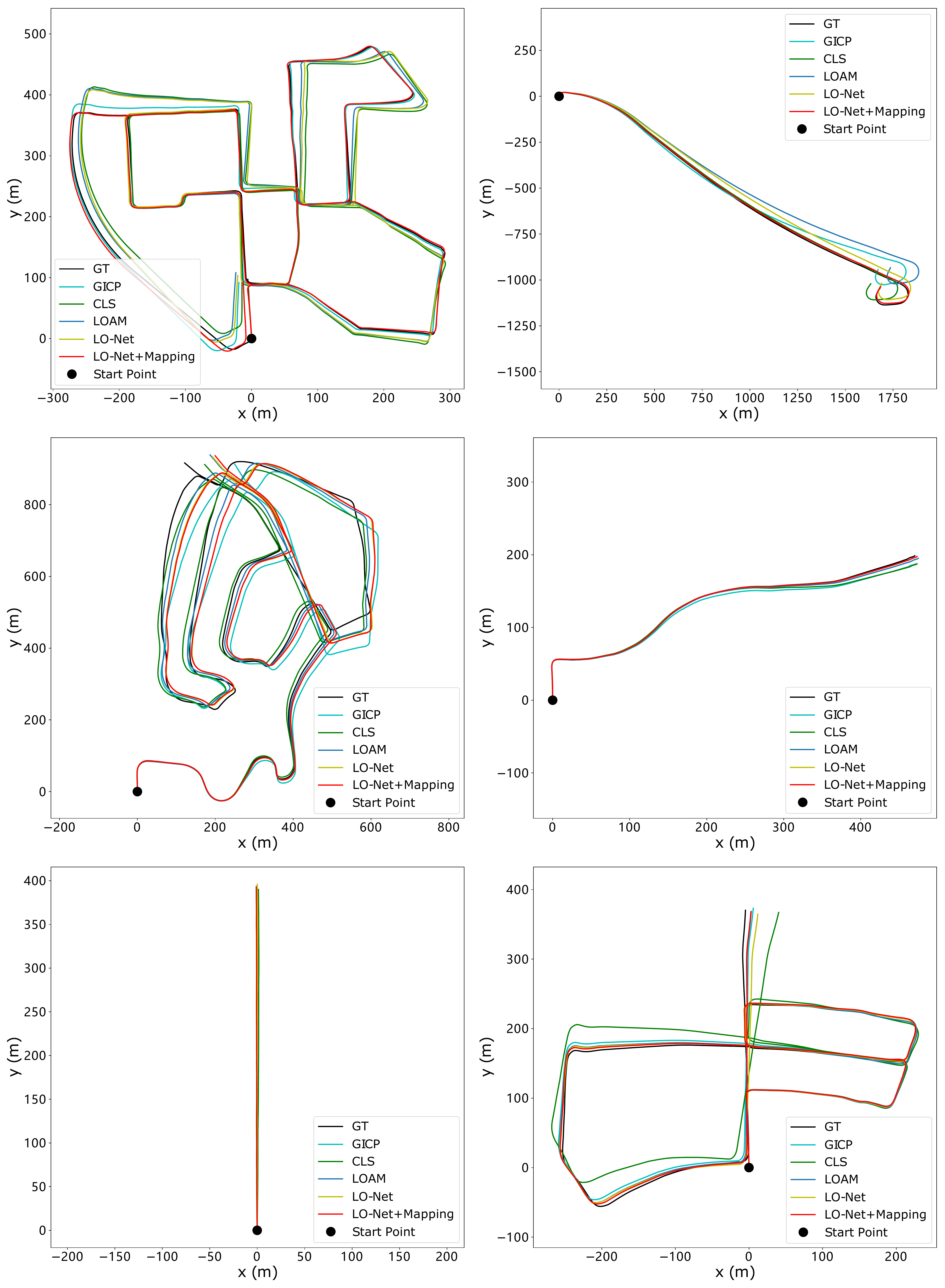}
	\caption{Trajectory plots of GICP~\cite{segal2009generalized}, CLS~\cite{velas2016collar}, LOAM~\cite{zhang2014loam}, LO-Net and LO-Net+Mapping on KITTI odometry Seq. 00-05.
		The ground truth trajectories are available. The results of ICP-po2pl are not shown as its large scale drift.}
	\label{fig:kitti1}
\end{figure*}

\begin{figure*}[!h]
	\centering
	\includegraphics[width=0.95\linewidth]{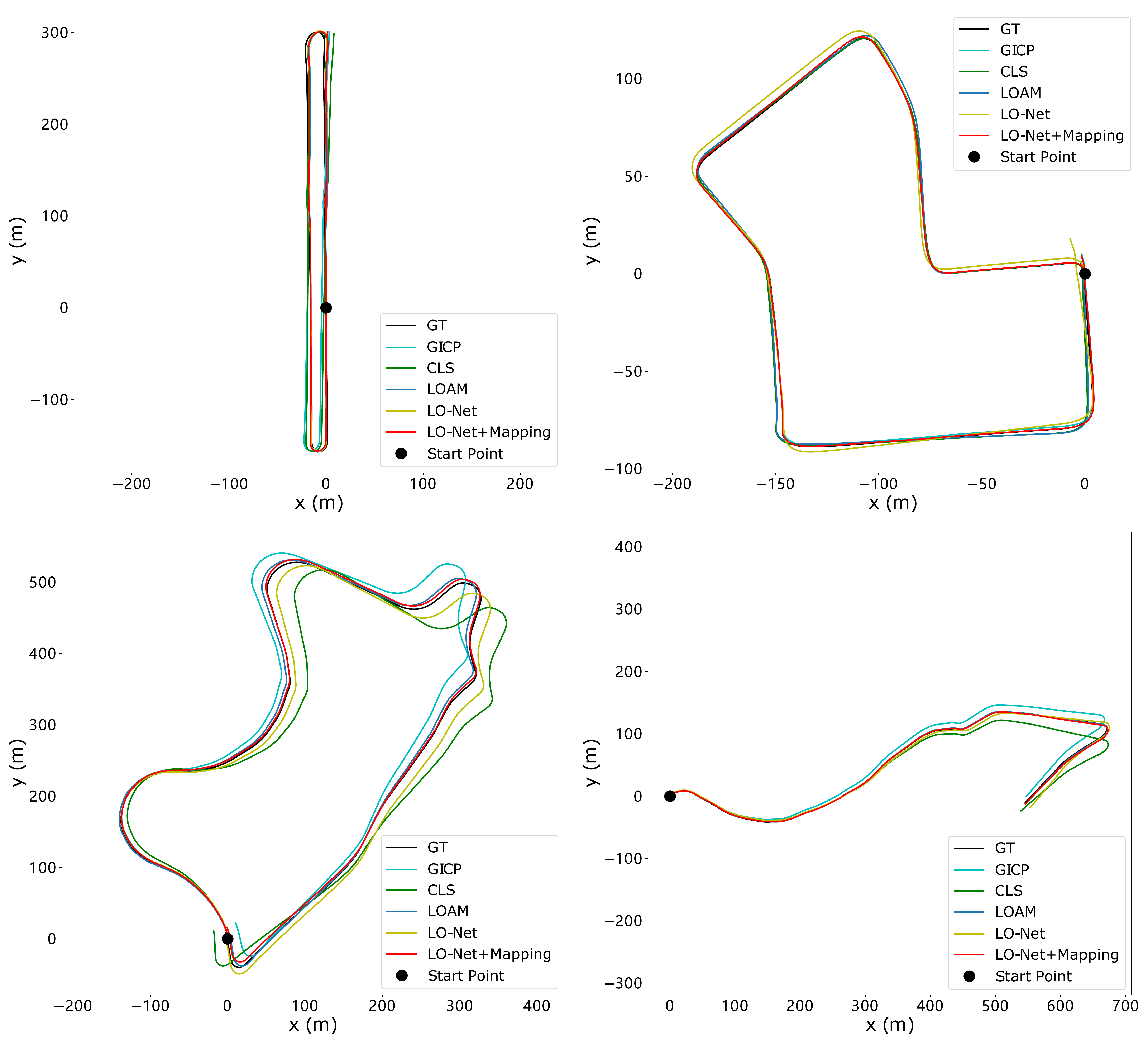}
	\caption{Trajectory plots of GICP, CLS, LOAM, LO-Net and LO-Net+Mapping on KITTI odometry Seq. 06, 07, 09, 10.		
		The ground truth trajectories are available. The results of Seq. 08 are shown in Section 4.2 of our paper.}
	\label{fig:kitti2}
\end{figure*}

\begin{figure*}[!h]
	\centering
	\includegraphics[width=0.9\linewidth]{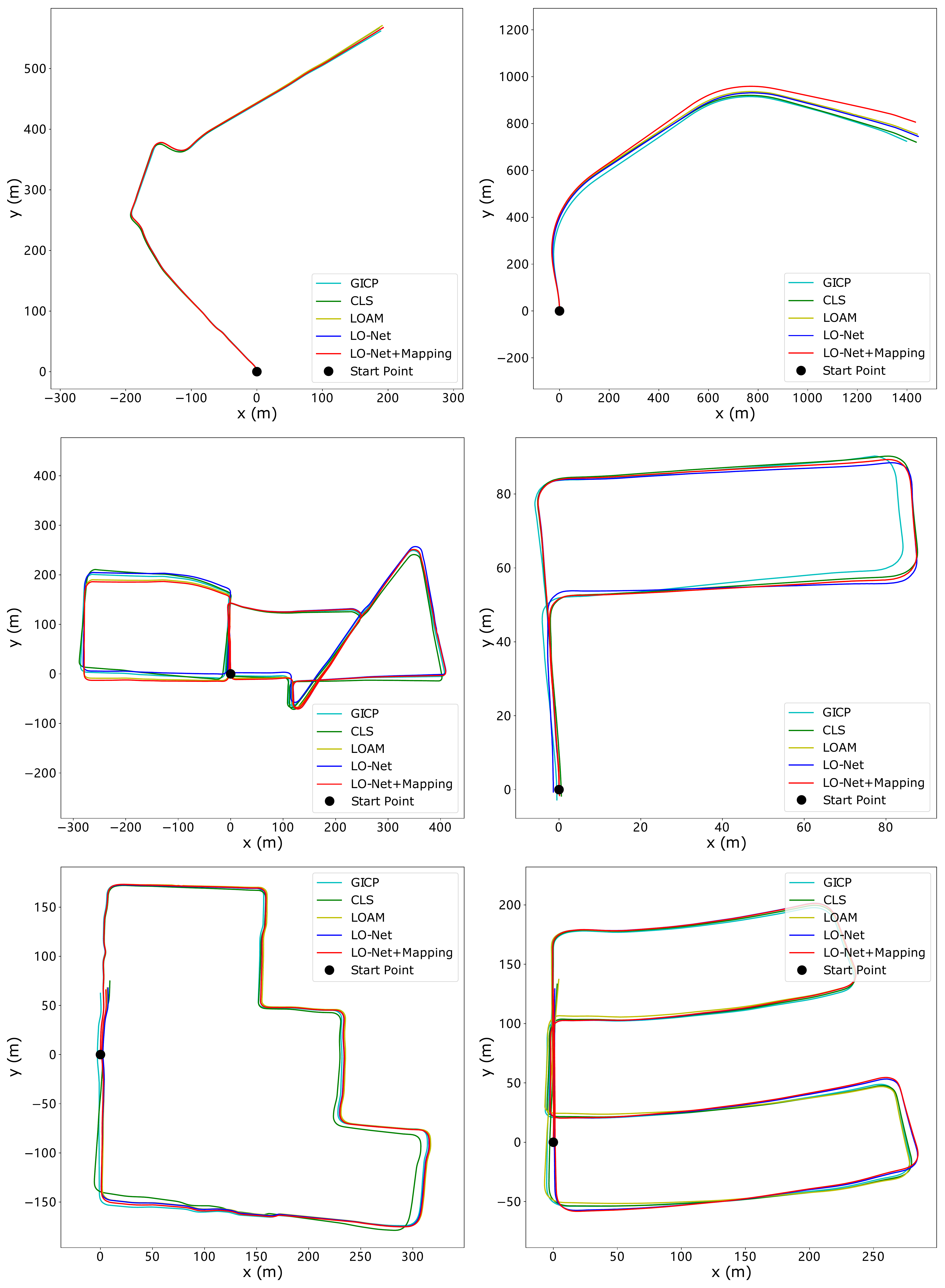}
	\caption{Trajectory plots of GICP, CLS, LOAM, LO-Net and LO-Net+Mapping on KITTI odometry Seq. 11-16.
		The ground truth trajectories are not available.}
	\label{fig:kitti3}
\end{figure*}

\begin{figure*}[!h]
	\centering
	\includegraphics[width=0.9\linewidth]{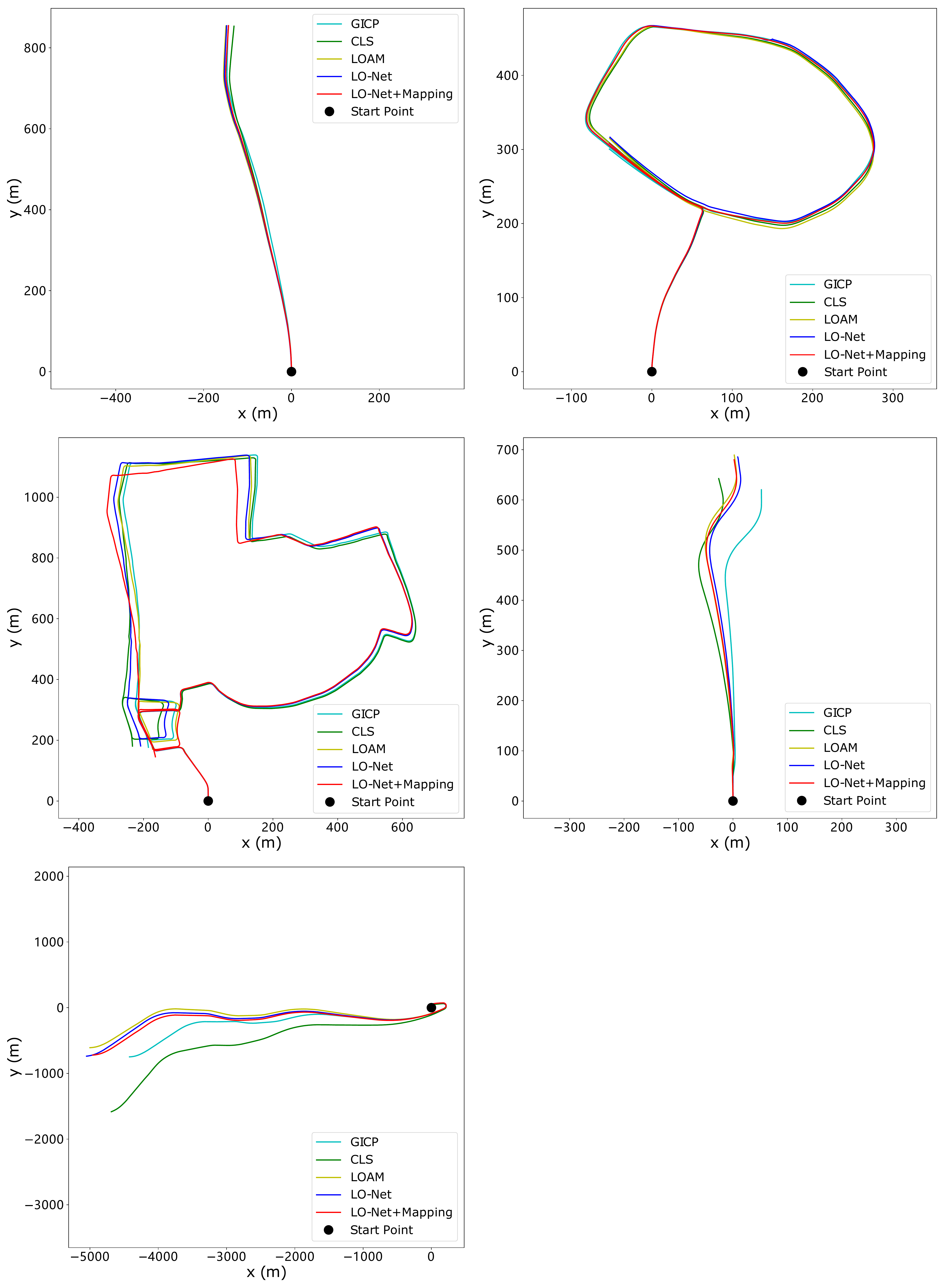}
	\caption{Trajectory plots of GICP, CLS, LOAM, LO-Net and LO-Net+Mapping on KITTI odometry Seq. 17-21.
		The ground truth trajectories are not available.}
	\label{fig:kitti4}
\end{figure*}

\begin{figure*}[!h]
	\centering
	\includegraphics[width=0.85\linewidth]{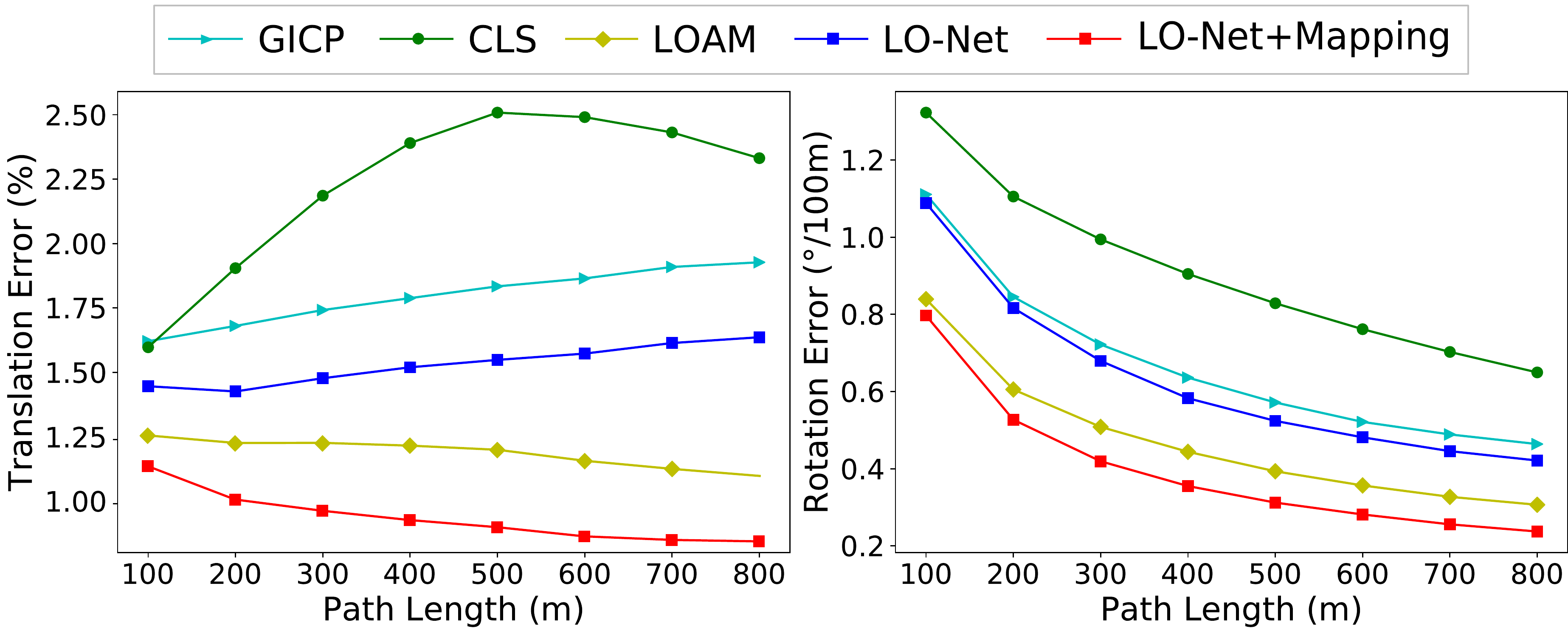}
	\caption{Average errors of translation and rotation with respect to path length intervals on KITTI Seq. 00-10.
		Our LO-Net+Mapping achieves the best performance among all the evaluated methods.}
	\label{fig:error_seg}
\end{figure*}

\begin{figure*}[!h]
	\centering
	\includegraphics[width=0.85\linewidth]{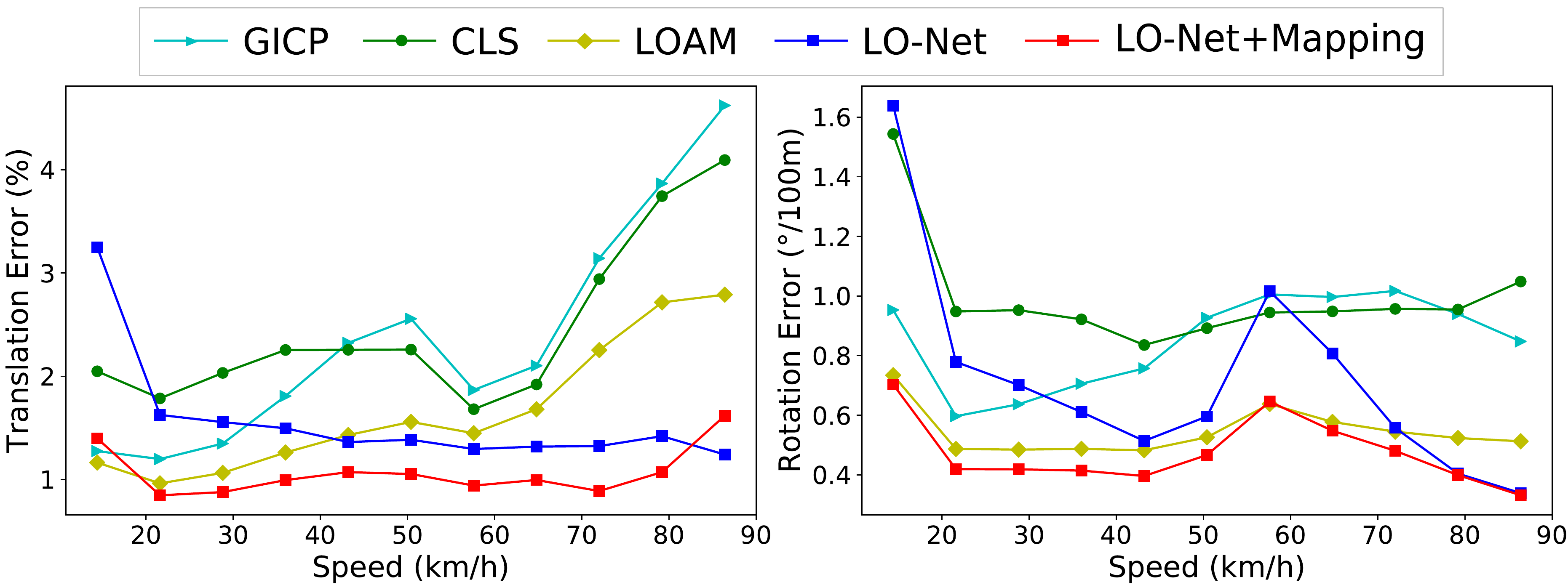}
	\caption{Average errors of translation and rotation with respect to driving speed on KITTI Seq. 00-10.
		Our LO-Net+Mapping achieves the best performance among all the evaluated methods.}
	\label{fig:error_speed}
\end{figure*}

\begin{figure*}[!h]
	\centering
	\includegraphics[width=0.48\linewidth]{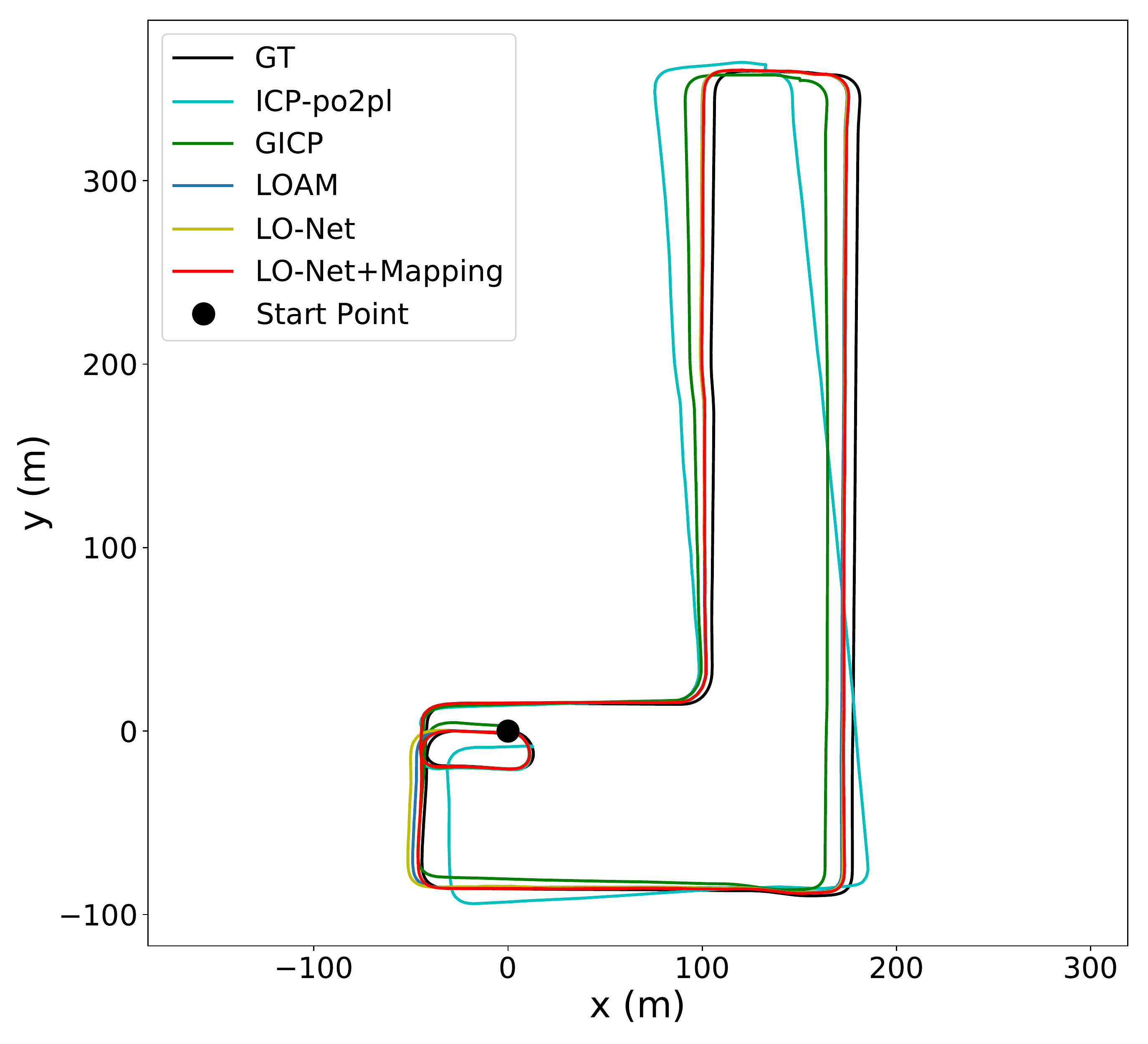}
	\includegraphics[width=0.48\linewidth]{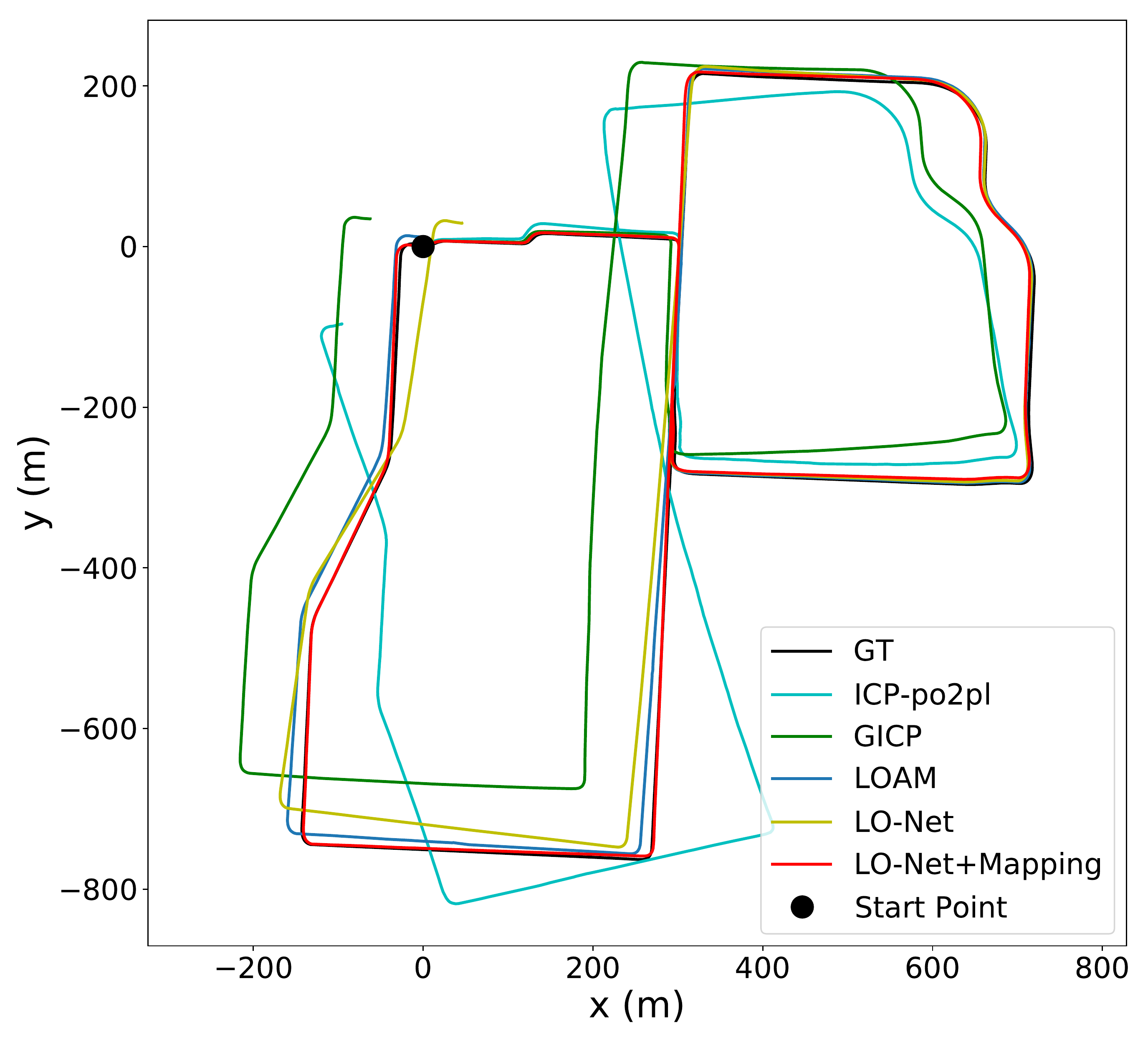}
	\caption{Trajectory plots of ICP-po2pl, GICP, LOAM, LO-Net and LO-Net+Mapping on Ford dataset.
		The ground truth trajectories are generated from IMU readings.
		The results of CLS are not shown due to its large scale drift.}
	\label{fig:ford}
\end{figure*}

In Figure~\ref{fig:kitti1} to~\ref{fig:error_speed}, we show the estimated trajectories and evaluation results of evaluated methods on KITTI odometry benchmark~\cite{geiger2012we}. 
In Figure~\ref{fig:kitti1} and~\ref{fig:kitti2}, the sequences are provided with ground truth. 
In Figure~\ref{fig:kitti3} and~\ref{fig:kitti4}, the ground truth is not available. 
The quantitative evaluation results on KITTI Seq. 00-10 are shown in Figure~\ref{fig:error_seg} and \ref{fig:error_speed}.
The estimated trajectories of evaluated methods on Ford dataset~\cite{pandey2011ford} are shown in Figures~\ref{fig:ford}. 
Evaluations on benchmark datasets demonstrate that our method outperforms existing ICP-based approaches 
and has similar accuracy with the state-of-the-art geometry-based approach.

\vspace*{\fill} 
\vspace*{\fill} 


\end{document}